\documentclass[runningheads]{llncs}
\usepackage[T1]{fontenc}
\usepackage{esvect}
\usepackage{subcaption}
\usepackage{caption}
\usepackage{array}
\usepackage{graphicx}
\usepackage{import}
\usepackage{url}
\usepackage{wrapfig}
\usepackage{hyperref}
\usepackage{multirow}
\usepackage{multicol}
\usepackage{siunitx, booktabs, tabularx}
\usepackage{tikz}
\usepackage{pgf}
\usepackage{pgfplots}
\usepackage{grffile}
\pgfplotsset{compat=1.17}
\usepgfplotslibrary{fillbetween}
\usetikzlibrary{arrows.meta,calc, decorations.pathreplacing}
\usepackage{verbatim}
\usepackage{amsmath}
\usepackage{amsfonts}
\usepackage{color}

\begin{document}
\title{Optimization is Not Enough: Why Problem Formulation Deserves Equal Attention}
%
%\titlerunning{Abbreviated paper title}
% If the paper title is too long for the running head, you can set
% an abbreviated paper title here
\titlerunning{Why Problem Formulation Deserves Equal Attention}
%\author{First Author\inst{1}\orcidID{0000-1111-2222-3333} \and
%econd Author\inst{2,3}\orcidID{1111-2222-3333-4444} \and
%Third Author\inst{3}\orcidID{2222--3333-4444-5555}}
%
%\author{
%Anonymous Authors
%}

\author{
Iván Olarte Rodríguez\inst{1}\orcidID{0009-0005-0748-9069} \and
Gokhan Serhat\inst{2}\orcidID{0000-0001-9928-0521} \and
Mariusz Bujny\inst{3}\orcidID{0000-0003-4058-3784} \and
Fabian Duddeck\inst{4}\orcidID{0000-0001-8077-5014} \and
Thomas Bäck\inst{1}\orcidID{0000-0001-6768-1478} \and
Elena Raponi\inst{1}\orcidID{0000-0001-6841-7409}
}

\authorrunning{Olarte Rodríguez et al.}

\institute{
LIACS, Leiden University, Leiden, The Netherlands \\
\email{\{i.olarte.rodriguez, t.h.w.baeck, e.raponi\}@liacs.leidenuniv.nl}
\and
Department of Mechanical Engineering, KU Leuven, Bruges, Belgium \\
\email{gokhan.serhat@kuleuven.be}
\and
NUMETO, Mikołów, Poland \\
\email{mariusz.bujny@numeto.eu}
\and
TUM School of Engineering and Design, Technical University of Munich, Munich, Germany \\
\email{duddeck@tum.de}
}
\maketitle

\begin{abstract}
Black-box optimization is increasingly used in engineering design problems where simulation-based evaluations are costly and gradients are unavailable.
In this context, the optimization community has largely analyzed algorithm performance in context-free setups, while not enough attention has been devoted to how problem formulation and domain knowledge may affect the optimization outcomes.
We address this gap through a case study in the topology optimization of laminated composite structures, formulated as a black-box optimization problem. Specifically, we consider the design of a cantilever beam under a volume constraint, intending to minimize compliance while optimizing both the structural topology and fiber orientations.
To assess the impact of problem formulation, we explicitly separate topology and material design variables and compare two strategies: a concurrent approach that optimizes all variables simultaneously without leveraging physical insight, and a sequential approach that optimizes variables of the same nature in stages.
Our results show that context-agnostic strategies consistently lead to suboptimal or non-physical designs. In contrast, the sequential strategy yields better-performing and more interpretable solutions. 
These findings underscore the value of incorporating, when available, domain knowledge into the optimization process and motivate the development of new black-box benchmarks that reward physically informed and context-aware optimization strategies.  
\end{abstract}

% Keywords section

% Introduction
\section{Introduction}
\label{sec:Introduction}
Optimization is a foundational tool in engineering design \cite{MartinsNing2021}. When coupled with a physically grounded model, it enables the systematic search for high-performance solutions using increasingly advanced algorithmic frameworks. However, the efficacy of such methods critically depends on the well-posedness and physical coherence of the optimization problem itself. As we argue in this work, optimization algorithms, regardless of their sophistication, do not inherently yield valid or practically useful solutions unless the problem is carefully formulated with appropriate modeling assumptions and structural considerations.

In the field of black-box optimization, considerable research has focused on improving optimization performance by either selecting an appropriate algorithm for the given optimization landscape (the Algorithm Selection Problem or ASP) \cite{Bischl2012,Kerschke2017}, or by tuning the algorithm’s configuration (the Algorithm Configuration Problem or ACP). A common approach used for both tasks is Exploratory Landscape Analysis (ELA), which provides descriptive features of the problem landscape based on a limited sampling of the objective function \cite{Mersmann2011,Munoz2015,Kerschke2016}. These features are then used to inform algorithm selection or configuration strategies, usually based on machine learning approaches, allowing a corresponding adaptation to the underlying structure of the problem. However, despite their relative success in synthetic benchmarks, such as the Black-Box Optimization Benchmarking (BBOB) test suite from the COmparing Continuous Optimizers (COCO) platform \cite{hansen_coco_2021}, these frameworks often become prohibitively expensive when applied to real-world problems due to limited evaluation budgets and computational constraints. Hence, few studies such as Long~et~al.\cite{long_generating_2024} have applied this approach to expensive real-world optimization problems. Nevertheless, unlike synthetic or abstract benchmark functions, real-world optimization problems often involve structured domain knowledge: variables are not merely arbitrary inputs, but correspond to meaningful physical quantities with known constraints and well-understood effects on the system's performance. This prior knowledge, which ranges from categorical distinctions to estimates of variable influence, necessitates a shift in perspective. Instead of treating variables as elements of a purely black-box landscape, domain-specific insights can be integrated into the pipeline to bypass the need for costly characterization.

This study investigates the role of problem formulation in topology and fiber path optimization of laminated composite structures. We consider a canonical yet illustrative benchmark: a linearly elastic cantilever beam subjected to a concentrated downward tip load. We introduce a benchmark with a new design space, parameterized using the Moving Morphable Components (MMC) method, which offers a compact and explicit geometric representation. To model spatially variable anisotropy of composites reinforced with curved fibers, we employ lamination parameters (LPs), which enable representing the fiber orientations in a compact form. This formulation enables a unified optimization of both the structural topology and the spatial distribution of fiber orientations, thus representing a novel integration of geometric and material design parameters.

Optimization is performed using both population-based and surrogate-based algorithms under limited function evaluation budgets. We compare the results obtained through sequential and concurrent optimization to illustrate that the latter can lead to infeasible or structurally invalid solutions even when using advanced optimizers.

This study emphasizes a key insight: the effectiveness of an optimization process is enhanced when the problem is well-defined and grounded in physical realism. By using a seemingly simple cantilever setup, we demonstrate how incorporating domain knowledge from the outset (via a sequential optimization approach) can improve results and guide algorithmic choices more effectively than considering a pure black-box approach. 

\section{Related Work}
\label{sec:related_work}

This section reviews key advances in optimization techniques relevant to composite structural design. We first cover black-box optimization methods used for problems with expensive or non-differentiable objectives. Then, we discuss progress in topology optimization, focusing on the MMC technique. Next, we examine fiber orientation optimization, including both lamination parameter and geometric formulations for variable-stiffness designs. Finally, we highlight integrated strategies that combine topology and fiber orientation optimization.

\subsection{Black-box Optimization}
Black-box Optimization (BBO) refers to the process of optimizing objective functions in which the analytic form of the objective is treated as unknown. Therefore, computation of local gradients is disregarded, making these methods useful when objective functions are expensive to evaluate and may exhibit noisy and non-smooth behavior. These scenarios commonly arise in engineering design, where objectives are computed via high-fidelity simulations or physical experiments. Classical derivative-free methods, such as the Nelder-Mead method, pattern search, or genetic algorithms, have traditionally been used for such problems, offering robustness but often requiring a large number of function evaluations.

More recently, model-based approaches such as Bayesian Optimization (BO) have gained popularity for their ability to balance exploration and exploitation through surrogate models---typically Gaussian Processes (GPs) for optimization tasks on continuous search spaces \cite{jones_efficient_1998}---and tailored candidate selection policies. The use of surrogate models makes these methods particularly effective for low-dimensional problems with expensive function evaluations.
To address the limitations of BO in high-dimensional or structured problems, several scalable variants have been proposed. The Trust Region Bayesian optimization (TuRBO) algorithm \cite{eriksson_scalable_2020} adapts trust regions dynamically to maintain sample efficiency in high dimensions. The Bayesian optimization with adaptively expanding subspaces (BAxUS) \cite{papenmeier_increasing_2023} incorporates low-dimensional subspace modeling via random embeddings to exploit problem sparsity, while the Heteroscedastic Evolutionary Bayesian optimization (HEBO) algorithm \cite{cowen-rivers_hebo_2022} combines input warping and hierarchical surrogate modeling for improved scalability. These methods exemplify the trend of integrating structure-exploiting strategies within BO to overcome the curse of dimensionality.

Population-based algorithms such as the Covariance Matrix Adaptation Evolution Strategy (CMA-ES) \cite{hansen_reducing_2003,hansen_completely_2001} have been applied in engineering design pipelines due to their simple configuration, low computational cost, and better handling of bifurcations and discontinuities of the optimization landscape than model-based approaches \cite{duddeck_multidisciplinary_2008}. However, they typically tend to be more exploitative and need larger evaluation budgets to converge to satisfactory solutions.  
\subsection{Topology Optimization}
\label{subsec:topology_optimization_review}
Topology Optimization (TO) is a key methodology for identifying optimal material layouts under defined loading conditions and structural performance criteria. Classical approaches such as the Solid Isotropic Material with Penalization (SIMP) \cite{sigmund_compliant_1997} and level-set methods \cite{allaire_structural_2004,wang_level_2003} have enabled automated design, though they can suffer from drawbacks such as blurred density distributions and complex numerical treatment, respectively. The so-called Bi-directional Evolutionary Structural Optimization (BESO) method \cite{huang_further_2010} helps improve definition by removing low-stress elements, but is limited to compliance-based objectives. To mitigate the curse of dimensionality and decouple the parametrization from a mesh, Guo et al.~\cite{guo2016explicit} proposed the Moving Morphable Components (MMC) approach, which uses geometric parameters as design variables to construct structures through movement, deformation, overlap, and deletion of members \cite{nozawa_topology_2025}. 
While effective, MMC restricts the design space based on the initial component set. Recent extensions by Bujny et al.~\cite{bujny_identification_2018} and Raponi et al.~\cite{raponi_kriging-assisted_2019} applied evolutionary and surrogate-based optimization strategies to MMC formulation, demonstrating the suitability of gradient-free methods for nonlinear problems such as vehicle crashworthiness, where traditional gradient-based techniques like the Method of Moving Asymptotes (MMA) \cite{svanberg_method_1987} may be less effective.
\subsection{Fiber Orientation Optimization}
Fiber-reinforced composites are widely used in performance-critical industries due to their excellent stiffness-to-weight properties. Optimization of fiber paths in these materials enhances structural performance by tailoring anisotropic stiffness attributes.
Early analytical methods aligned fibers with principal stress directions \cite{pedersen_optimal_1989,pedersen_thickness_1991}. These studies were later refined through energy-based formulations \cite{luo_optimal_1998}, highlighting conditions where principal stress alignment may not yield a unique optimal orientation. 

More recent numerical approaches, especially those using Finite Element Analysis (FEA), enable optimized fiber orientation at the element level through methods such as Continuous Fiber Angle Optimization (CFAO) \cite{petrovic_orthotropic_2018,setoodeh_design_2009}, supporting smooth, spatially varying fiber distributions. In contrast, LP-based methods \cite{grenestedt_layup_1989,miki_optimum_1993}, which reduce design dimensionality, are mostly limited to constant-stiffness laminates. The Lamination Parameter Interpolation Method (LPIM) \cite{serhat_multi-objective_2019,serhat_unifying_2020} allows meshless and variable-stiffness designs, with extensions to handle geometric discontinuities \cite{shafighfard_additive_2021,shafighfard_design_2019,sheikhi_design_2024} and manufacturability constraints \cite{guo_design_2022,rafiei_anamagh_eigenfrequency_2024}.

Other techniques use parametric representations like polynomials \cite{honda_multi-objective_2013}, splines \cite{nagendra_optimization_1995}, streamlines \cite{tosh_design_2000}, and level-sets \cite{tian_parametric_2022}, enabling closed-form fiber paths that offer better curvature and spacing control \cite{shimoda_numerical_2022,tian_optimization_2023,zhang_non-uniform_2024}.
%To reduce computational costs, some approaches use surrogate models, including truss analogies \cite{zhang_nodal-based_2023} and stress-based sampling with adjoint optimization \cite{sun_more_2023}.

In terms of optimization strategies, gradient-based methods are common but prone to converging to local minima \cite{shen_orientation_2020}, while hybrid and derivative-free approaches (e.g., particle swarm optimization) \cite{khandar_shahabad_advanced_2023} have proven effective for both LP and geometric formulations. On the other hand, meta-heuristics and genetic algorithms, as in \cite{shafei_optimizing_2024}, have been used for problem formulations where the fiber angle parameterization (such as NURBS) does not allow computing sensitivities. 

%Overall, the field has evolved from simple rule-based orientation strategies to sophisticated, computationally tractable optimization frameworks that address not only performance but also manufacturability and robustness. Yet, challenges remain in integrating these techniques with broader structural objectives and extending them to large-scale and 3D composite systems.
%
%In parallel, topology optimization (TO) has been employed to reduce material use and enhance structural efficiency. The integration of TO with fiber orientation design has led to two main approaches: discrete material optimization (DMO), which assigns fiber angles from a predefined set, and continuous fiber angle optimization (CFAO), which allows smoothly varying fiber orientations across the domain. Both approaches directly manipulate local fiber angles and are referred to as explicit methods, with CFAO offering advantages in continuity and design flexibility.
%

\subsection{Mixed topology and fiber orientation optimization techniques}
TO and fiber orientation optimization have traditionally been addressed separately in the design of fiber-reinforced composites. However, recent research has increasingly focused on their integration to fully exploit the anisotropic behavior of composites. For more information about this topic, the reader is advised to check \cite{nozawa_topology_2025}.

The early work primarily adopted sequential optimization strategies, in which structural topology and fiber orientations were optimized in two distinct stages. This decoupling simplifies problem formulation and improves numerical stability. For instance, Peeters et al.~\cite{peeters_combining_2015}, based on an adaptation of the SIMP combined with LPs, first determined the optimal topology and then the LPs before constructing corresponding fiber paths. Other methods, such as those based on cellular automata \cite{setoodeh_combined_2005}, aimed to enhance computational efficiency and robustness.

In contrast, concurrent optimization approaches have gained popularity for their potential to achieve superior structural performance. Techniques combining SIMP with fiber orientation variables \cite{schmidt_structural_2020}, or those based on homogenization theory \cite{kim_topology_2020}, allow concomitant optimization of material distribution and anisotropy. One particular approach is the one developed by Sun~et al.~\cite{sun_structural_2022}, wherein the topology and the fiber orientation were optimized by using a joint formulation of MMC by defining each component with a fixed fiber orientation. 

However, most of the aforementioned methods typically involve a large number of design variables, especially when both topology and orientation fields are defined at fine resolution. To address this issue, we propose a hybrid approach that combines MMCs for TO with LPs for fiber steering. This formulation reduces the dimensionality of the design space and enables gradient-free concurrent optimization, which we benchmark in this paper against sequential and context-aware strategies.

\section{Problem Formulation}
\label{sec:Problem Formulation}
This section outlines the formulation of the TO problem for laminated composite structures, wherein the material distribution and the local stiffness properties are simultaneously optimized. The domain is described using a level-set-based geometry representation, while the anisotropic behavior of laminated composites is represented through LPs derived from classical laminate theory. A key aspect of the formulation is the integration of the Lamination Parameter Interpolation Method (LPIM), which enables continuous and physically feasible variation of stiffness properties across the structure. The resulting optimization problem seeks to determine both the optimal layout of structural material and the spatial distribution of LPs such that structural performance objectives are met under manufacturing and physical constraints. 
%The parameterization strategies, interpolation schemes, and finite element modeling assumptions used to define this problem are detailed in the following subsections.

% Optional continuation for FEA integration
\subsection{Design Domain and Finite Element Formulation}
\label{subsec:design_domain}
\begin{figure}[htbp!]
    \centering
    \begin{tikzpicture}[scale=1.2]
% All font is small sized
\tikzset{every node/.style={font=\small}}
% Parameters
\def\L{4}
\def\H{2}
\def\nx{120} % number of divisions in x
\def\ny{60}  % number of divisions in y
\def\dx{\L/\nx}
\def\dy{\H/\ny}

% Beam rectangle
\draw[very thick] (0,0) rectangle (\L,\H);

% Mesh
\foreach \i in {1,...,\nx} {
    \draw[gray!50] ({\i*\dx},0) -- ({\i*\dx},\H);
}
\foreach \j in {1,...,\ny} {
    \draw[gray!50] (0,{\j*\dy}) -- (\L,{\j*\dy});
}

% Fixed boundary on the left
\foreach \y in {0.2, 0.5, ..., 1.8} {
    \draw[thick] (-0.2,\y) -- (0,\y+0.2);
}
\node at (0,2.23) {Fixed Edge};

% Load arrow on right middle
\draw[very thick,->,red] (\L,1) -- (\L,0.3);
\node[right, red] at (\L+0.1,0.65) {${|\overrightarrow{F}|}=0.10$};
\node at (\L+0.8,1.3) {Load};

% Labels
\draw[<->, thin] (0,-0.5) -- (\L,-0.5) node[midway, below] {$L$ (100 elements)};
% Plotting the vertical lines
\draw[-, very thin] (0,0) -- (0,-0.65);
\draw[-, very thin] (\L,0) -- (\L,-0.65);

\draw[<->, thin] (-0.5,0) -- (-0.5,\H)
    node[midway, left, align=center] {$H$ \\ (50 elements)};

% Plotting the horizontal lines
\draw[-, very thin] (0,0) -- (-0.65,0.0);
\draw[-, very thin] (0,\H) -- (-0.65,\H);

\end{tikzpicture}
    \caption{Schematics of the cantilever beam test case. The depicted elements have a length of 1 in both $x$ and $y$ directions.}
    \label{fig:Cantilever_beam_representation}
\end{figure}
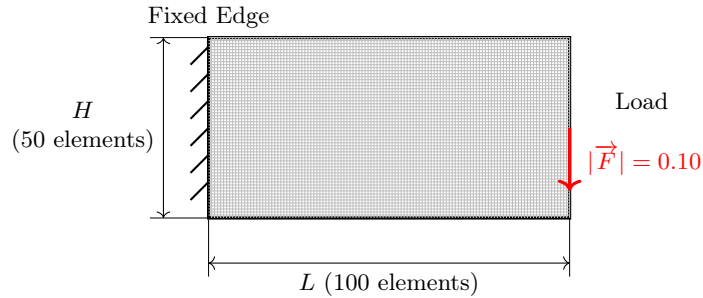

The structural domain investigated in this work comprises a cantilever rectangular laminate plate subjected to a concentrated downward force applied at the midpoint of the right edge, as depicted in Figure~\ref{fig:Cantilever_beam_representation}. The domain is discretized using a finite element mesh consisting of $100 \times 50$ elements, thereby preserving a $2$:$1$ aspect ratio aligned with established benchmarks \cite{bujny_identification_2018,guo2016explicit,raponi_kriging-assisted_2019}. This discretization governs the spatial resolution of the design field for subsequent TO.

The mechanical response of the composite laminate is simulated through linear elastic theory. Specifically, 4-node rectangular isoparametric Reissner-Mindlin shell elements with unit thickness are employed. The in-plane stiffness matrix of each element is computed using Equation \eqref{eq:in_plane_stiffness} defined in Section~\ref{subsubsec:LPIM}, according to the definition of LP distribution, whose formulation is detailed in the same section.
An in-house finite element solver was utilized to perform the simulations.

\subsection{Parameterization}
\label{subsec:Parameterization}
\subsubsection{Moving Morphable Components.}
\label{subsubsec:MMC}
Let \( D \subset \mathbb{R}^2 \) be a bounded design domain (as in Figure \ref{fig:Cantilever_beam_representation}) that contains all admissible layouts. Denote by \( \Omega \subset D \) the subregion occupied by the material. The global Level-Set Function (LSF), \( \Phi: D \rightarrow \mathbb{R} \), characterizes the material distribution via the following sign convention:
\[
\begin{cases}
\Phi(\mathbf{q}) > 0, & \mathbf{q} \in \Omega, \\
\Phi(\mathbf{q}) = 0, & \mathbf{q} \in \partial \Omega, \\
\Phi(\mathbf{q}) < 0, & \mathbf{q} \in D \setminus \Omega,
\end{cases}
\]
where \( \partial \Omega \) denotes the material-void interface. The zero-level set of \( \Phi \) thus defines the interface between distinct material phases, ensuring a crisp boundary representation without intermediate densities at any point \( \mathbf{q} = (x, y)^\top \in D \).

The global LSF \( \Phi \) is constructed as a composition of local level-set functions \( \phi_i: D \rightarrow \mathbb{R} \), each associated with an individual, deformable beam-like structural component, referred to as MMC. 
% Each local function \( \phi_i \) defines a subdomain \( \Omega_i \subset D \) through the relation:
% \begin{equation}
% \begin{cases}
% \phi_i(\mathbf{q}) > 0, & \mathbf{q} \in \Omega_i, \\
% \phi_i(\mathbf{q}) = 0, & \mathbf{q} \in \partial \Omega_i, \\
% \phi_i(\mathbf{q}) < 0, & \mathbf{q} \in D \setminus \Omega_i,
% \end{cases}
% \label{eq:LSF_definition}
% \end{equation}
% where \( \Omega_i \) is the region occupied by the \( i \)-th component. If the design is composed of \( N \) such components, the total material domain is given by:
% \begin{equation}
% \Omega = \bigcup_{i=1}^N \Omega_i.
% \label{eq:material_union}
% \end{equation}

For a two-dimensional setting, each component is represented by a local level-set function of the following form:
\begin{align}
\phi_i(\mathbf{q}) = - \Bigg[ 
& \left( \frac{
\cos \theta_i (x - x_{c,i}) + \sin \theta_i (y - y_{c,i})
}{l_i / 2} \right)^m \nonumber \\
& + \left( \frac{
- \sin \theta_i (x - x_{c,i}) + \cos \theta_i (y - y_{c,i})
}{t_i / 2} \right)^m 
- 1 \Bigg],
\label{eq:individual_LSF}
\end{align}
where \( (x_{c,i}, y_{c,i}) \in D \) denotes the center of the \( i \)-th component, \( l_i \) and \( t_i \) are its length and thickness, respectively, and \( \theta_i \) is its in-plane orientation angle. The exponent \(m\) is chosen as a relatively large even integer, set to \( m=6 \) in this work as commonly adopted in the literature \cite{bujny_identification_2018,guo2016explicit,raponi_kriging-assisted_2019}.

\begin{figure}[htbp]
\centering
\begin{subfigure}{0.45\textwidth}
    \centering
    \begin{tikzpicture}[scale=1, every node/.style={font=\small}]

% Parameters
\def\l{4}       % length of straight part
\def\t{1}       % diameter (t)
\def\r{0.5}     % radius = t/2
\def\beta{30}  % rotation angle

% Capsule center
\coordinate (C) at (0,0);

% Draw capsule shape
\begin{scope}[rotate=\beta]
    % Capsule endpoints
    \coordinate (A) at (-\l/2,0);
    \coordinate (B) at (\l/2,0);
    \coordinate (E) at (\l/2+0.5,0);
    
    % Outer capsule
    \draw[thick] 
        ({-\l/2 + \r}, \r)
        arc[start angle=90, end angle=270, radius=\r] --
      ({\l/2 - \r}, -\r)
        arc[start angle=90, end angle=270, radius=-\r] --
      cycle;

    % Dashed centerline
    \draw[dashed, , thin] (A) -- (E);
    
    % Length label
    \draw[<->, very thin] ($(A)+(0,1.4*\r)$) -- node[above] {$l$} ($(B)+(0,1.4*\r)$);
    % Vertical lines to denote the measurement
    \draw[-, very thin] ($(A)$) -- ($(A)+(0,1.58*\r)$);
    \draw[-, very thin] ($(B)$) -- ($(B)+(0,1.58*\r)$);
    
    % Height label (t)
    \draw[<->, very thin] ($(A)+(-1.2*\r,\r)$) -- node[left] {$t$} ($(A)+(-1.2*\r,-\r)$);
    % horizontal lines to denote the measurement
    \draw[-, very thin] ($(A)+(-1.2*\r,\r)$) -- ($(A)+(\r,\r)$) ;
    \draw[-, very thin] ($(A)+(-1.2*\r,-\r)$) -- ($(A)+(\r,-\r)$) ;
    % Center point
    
    \fill (C) circle (1pt);
    \node [rotate=\beta, above left] at (C)  {$(x_{c}, y_{c})$};

\end{scope}

% Angle theta arc
% Dashed centerline
\coordinate (D) at (\l/2+0.5,0);
\draw[dashed, , thin] (C) -- ++(D);
\draw[<->, very thin] (0,0) ++(D) arc[start angle=0, end angle=\beta, radius=\l/2+0.5];
\node at ($(0,0)+(\l/2+0.15,0.55)$) {$\theta$};

\end{tikzpicture}
    \caption{Parameterization of an individual MMC.}
    \label{fig:individual_param_mmc}
\end{subfigure}
\hfill
\begin{subfigure}{0.45\textwidth}
    \centering
    \includegraphics[width=\textwidth]{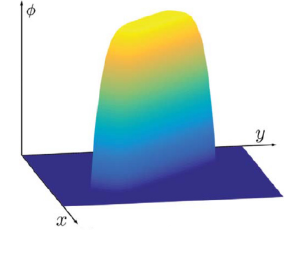}
    \caption{Corresponding LSF of an individual component.}
    \label{fig:sole_lsf}
\end{subfigure}
\caption{Parametrization and LSF of a single MMC. Adapted from \cite{raponi_kriging-assisted_2019}.}
\label{fig:Level-set_function_individual}
\end{figure}
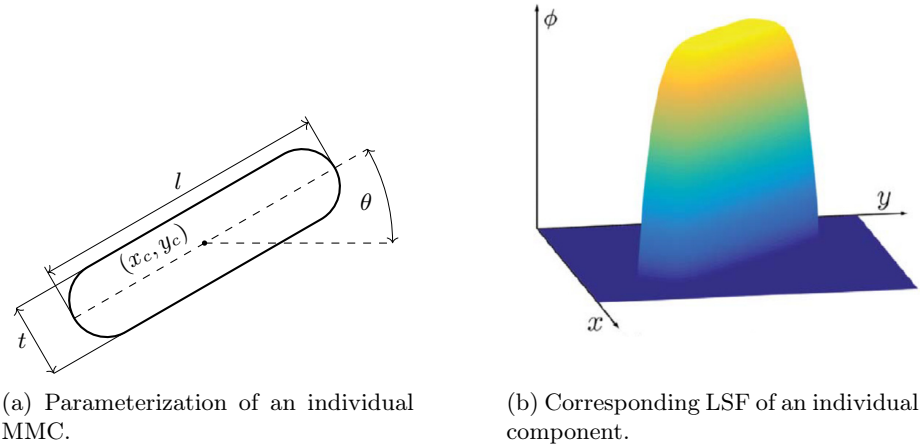

Each component is parameterized by the vector \( (x_{c,i}, y_{c,i}, \theta_i, l_i, t_i) \), providing a compact representation of its geometry. Figure \ref{fig:Level-set_function_individual} provides a visualization of the MMC's parametrization and the corresponding to local LSF, respectively. 

\subsubsection{Lamination Parameters.}
\label{subsubsec:LPs}
The LP-approach provides a compact means of representing the stiffness characteristics of laminated composites \cite{tsai1980introduction}.
LPs serve as design variables, replacing explicit properties such as the number of plies, their thicknesses, and fiber orientations.
One major advantage of this formulation is its tendency to yield convex design spaces, which typically eliminates issues related to local optima \cite{grenestedt1993layup,serhat_multi-objective_2019}.
% Figure~\ref{fig:curved_panel} illustrates the schematic of the curved composite panel used in this study. The local coordinate system is defined such that the $x$-axis follows the panel's longitudinal direction and rotates with it, the $y$-axis lies along the width and remains fixed, while the $z$-axis points through the panel's thickness and is normal to its surface.

% %\begin{figure}[htbp!]
% \begin{wrapfigure}{l}{5.25cm}
%     \centering
%     \includegraphics[width=0.35\textwidth]{./Figures/laminate.pdf}
%     \caption{Curved laminated composite panel.}
%     \label{fig:curved_panel}
% \end{wrapfigure}

% The lamination parameters approach provides a compact means of representing the stiffness characteristics of laminated composites~\cite{tsai1980introduction}. This method expresses the overall stiffness matrices in terms of material invariants and lamination parameters. While the material invariants are constant and depend solely on the material properties $(E_1,E_2,G_{12},\nu_{12})$, the lamination parameters serve as design variables, replacing explicit details such as the number of plies, their thicknesses, and fiber orientations. One major advantage of this formulation is its tendency to yield convex design spaces, which typically eliminates issues related to local optima \cite{grenestedt1993layup,serhat_multi-objective_2019}.

According to classical laminated plate theory, the constitutive relations for a laminated composite plate are given as \cite{gurdal1999design}:
\begin{equation}
	\begin{Bmatrix}
	N_x \\
	N_y \\
	N_{xy} \\
	M_x \\
	M_y \\
	M_{xy}
	\end{Bmatrix}
	=
	\begin{bmatrix}
	\mathbf{A}^{0} & \mathbf{B}^{0} \\
	\mathbf{B}^{0} & \mathbf{D}^{0}
	\end{bmatrix}
	\begin{Bmatrix}
	\varepsilon_x^0 \\
	\varepsilon_y^0 \\
	\varepsilon_{xy}^0 \\
	\kappa_x \\
	\kappa_y \\
	\kappa_{xy}
	\end{Bmatrix},
\label{eq:constitutive_relation_composite}
\end{equation}
where $N_x$, $N_y$, and $N_{xy}$ are the in-plane force resultants per unit width, and $M_x$, $M_y$, and $M_{xy}$ are the corresponding bending moment resultants. The terms $\varepsilon_x^0$, $\varepsilon_y^0$, and $\varepsilon_{xy}^0$ denote mid-plane strains, while $\kappa_x$, $\kappa_y$, and $\kappa_{xy}$ represent curvatures. The matrices $\mathbf{A}^{0}$, $\mathbf{B}^{0}$, and $\mathbf{D}^{0}$ correspond to in-plane stiffness, bending-extension coupling, and bending stiffness, respectively. These matrices can be expressed in terms of the LPs and material invariants.

For symmetric and balanced laminates, the normalized in-plane stiffness matrix can be represented using two LPs ($V_1$ and $V_3$):
\begin{equation}
\frac{\mathbf{A}^{0}}{h} = 
\begin{bmatrix}
U_1 & U_4 & 0 \\
U_4 & U_1 & 0 \\
0 & 0 & U_5
\end{bmatrix}
+
\begin{bmatrix}
U_2 & 0 & 0 \\
0 & -U_2 & 0 \\
0 & 0 & 0
\end{bmatrix} V_1
+
\begin{bmatrix}
U_3 & -U_3 & 0 \\
-U_3 & U_3 & 0 \\
0 & 0 & -U_3
\end{bmatrix} V_3
.
\label{eq:in_plane_stiffness}
\end{equation}
$V_1$ and $V_3$ are defined as:
\begin{equation}
\begin{Bmatrix}
V_1 \\
V_3
\end{Bmatrix}
=
\frac{1}{h} \sum_{k=1}^{N_l} \varsigma_k
\begin{Bmatrix}
\cos(2\alpha_k) \\
\cos(4\alpha_k)
\end{Bmatrix},
\end{equation}

where $N_l$ is the number of layers, $\varsigma_k$ is the thickness of the $k$-th ply, $\alpha_k$ is its fiber orientation angle, $h$ is the total thickness of the laminate, and $U_j$ are the material invariants computed from the material properties (for further details, see~\cite{gurdal1999design}).

% The material invariants $U_j$ are computed as:
% \begin{equation}
% \begin{Bmatrix}
% U_1 \\
% U_2 \\
% U_3 \\
% U_4 \\
% U_5
% \end{Bmatrix}
% =
% \begin{bmatrix}
% \frac{3}{8} & \frac{3}{8} & \frac{1}{4} & \frac{1}{2} \\
% \frac{1}{2} & -\frac{1}{2} & 0 & 0 \\
% \frac{1}{8} & \frac{1}{8} & -\frac{1}{4} & -\frac{1}{2} \\
% \frac{1}{8} & \frac{1}{8} & \frac{3}{4} & -\frac{1}{2} \\
% \frac{1}{8} & \frac{1}{8} & -\frac{1}{4} & \frac{1}{2}
% \end{bmatrix}
% \begin{Bmatrix}
% Q_{11} \\
% Q_{22} \\
% Q_{12} \\
% Q_{66}
% \end{Bmatrix}.
% \label{eq:invariant_to_reduced_stiffness}
% \end{equation}

% %
% The reduced stiffness terms $Q_{kl}$ are calculated from material properties as:
% \begin{equation}
% \begin{Bmatrix}
% Q_{11} \\
% Q_{22} \\
% Q_{12} \\
% Q_{66}
% \end{Bmatrix}
% =
% \begin{Bmatrix}
% \frac{E_1}{\gamma} \\
% \frac{E_2}{\gamma} \\
% \frac{\nu_{12} E_2}{\gamma} \\
% G_{12}
% \end{Bmatrix}
% \quad \text{with} \quad \gamma = 1 - \nu_{12} \nu_{21}, \quad \nu_{21} = \frac{\nu_{12} E_2}{E_1}.
% \label{eq:reduced_stiffness_relations}
% \end{equation}
%
For symmetric laminates, the coupling matrix $\mathbf{B}^{0}$ vanishes. Furthermore, in laminates with many uniformly distributed plies, the bending stiffness matrix $\mathbf{D}^{0}$ becomes approximately orthotropic and satisfies the relation $\mathbf{D}^{0} = \mathbf{A}^{0} h^2 / 12$ \cite{fukunaga_stiffness_1991}. This approximation allows $\mathbf{D}^{0}$ to be expressed in terms of the LPs.

\subsubsection{Lamination Parameter Interpolation Method.}
\label{subsubsec:LPIM}
%
% The lamination parameter interpolation method enables the spatial variation of fiber orientations across a composite structure while ensuring manufacturability and structural feasibility. 
The LPIM enables the spatial variation of fiber orientations across a composite structure, while ensuring feasibility through a continuous design space derived from Miki’s diagram \cite{miki_optimum_1993}.
Although this diagram provides discrete feasible stiffness configurations for individual panels or plates, a continuous variable stiffness (VS) description requires each element to discretize the domain \( D \subset \mathbb{R}^2 \) to have an independent LP description. 
Following the approach outlined by Serhat and Basdogan~\cite{serhat_lamination_2019}, the interpolation procedure consists of the following steps: (1) generating the lamination parameter distribution, (2) computing the corresponding fiber angle field, and (3) reconstructing the fiber paths. In the following, we restrict our attention to the first step, as it is the most relevant to the optimization problem definition, given that it determines the optimal values of the design variables in our formulation.
% \begin{enumerate}
%     \item Lamination Parameter Distribution Generation;
%     \item Computation of the fiber angle distribution;
%     \item Reconstruction of the fiber paths.
% \end{enumerate}

% Only the first two steps are addressed herein for brevity, since the third step implies solving an additional design problem whose results depend on a criterion used \cite{blom_optimization_2010} and the fiber angle distribution is enough to visually compare the performance of Optimizers.

\begin{figure}[htbp]
    \centering
    \begin{subfigure}[t]{0.45\textwidth}
    \centering
    	\resizebox{1.1\linewidth}{!}{\begin{tikzpicture}[scale=0.9]
\begin{axis}[
    axis equal image,
    width=8cm,
    xlabel={$V_1$},
    ylabel={$V_3$},
    xmin=-1.00001, xmax=1.00001,
    ymin=-1.00001, ymax=1.00001,
    xtick={-1, -0.5, 0, 0.5, 1},
    ytick={-1, -0.5, 0, 0.5, 1},
    domain=-1:1,
    samples=200,
    axis y line*=left,
    axis x line*=bottom,
    enlargelimits=false,
    clip=false,
    colormap name=yellowred,
]

% Parabola with color gradient
\addplot[name path=lower, blue, thick, forget plot, domain=-1:0, samples=200] {2*x^2 - 1};
\addplot[name path=lower, green, thick, forget plot, domain=0:1, samples=200] {2*x^2 - 1};
\addplot[name path=lower,green!25!blue!75, thick, forget plot, domain=-0.5:0, samples=200] {2*(x/(2*0.25-1))^2 - 1};
\addplot[
name path=lower,green!50!blue!50, thick, forget plot, domain=-1:1, samples=200,
    parametric
] 
({0}, {x});  % x(t) = 1, y(t) = t
\addplot[name path=lower, green!75!blue!25, thick, forget plot, domain=0:0.5, samples=200] {2*(x/(2*0.25-1))^2 - 1};

% Text labels
\node[blue] at (axis cs:-0.78,1.025) {\tiny $R_l = 1.00$};
\node[blue] at (axis cs:-0.78,0.925) {\tiny $R_r = 0.00$};
\node[green!25!blue!75] at (axis cs:-0.245,1.025) {\tiny $R_l = 0.75$};
\node[green!25!blue!75] at (axis cs:-0.245,0.925) {\tiny $R_r = 0.25$};
\node[green!50!blue!50] at (axis cs:0.2,1.025) {\tiny $R_l = 0.50$};
\node[green!50!blue!50] at (axis cs:0.2,0.925) {\tiny $R_r = 0.50$};
\node[green!75!blue!25] at (axis cs:0.72,1.025) {\tiny $R_l = 0.25$};
\node[green!75!blue!25] at (axis cs:0.72,0.925) {\tiny $R_r = 0.75$};
\node[green] at (axis cs:1.2,1.025) {\tiny $R_l = 0.00$};
\node[green] at (axis cs:1.2,0.925) {\tiny $R_r = 1.00$};
\end{axis}
\end{tikzpicture}}
    	\caption{Several curves of design points for different volumetric ratios.}
    	\label{fig:Miki_Lamination_Diagram_Interpolation_2}
    \end{subfigure}
     \hfill % Space between the subfigures
   \begin{subfigure}[t]{0.45\textwidth}
   	    \centering
    	\resizebox{1.1\linewidth}{!}{\begin{tikzpicture}[scale=0.9]
\begin{axis}[
    axis equal image,
    width=8cm,
    xlabel={$V_1$},
    ylabel={$V_3$},
    xmin=-1.00001, xmax=1.00001,
    ymin=-1.00001, ymax=1.00001,
    xtick={-1, -0.5, 0, 0.5, 1},
    ytick={-1, -0.5, 0, 0.5, 1},
    domain=-1:1,
    samples=200,
    axis y line*=left,
    axis x line*=bottom,
    enlargelimits=false,
    clip=false,
    colormap name=yellowred,
]

% Custom colormap
\pgfplotsset{
  colormap={yellowred}{
    rgb(0cm)=(1,1,0);
    rgb(1cm)=(1,0,0)
  }
}

% Parabola with color gradient
\addplot[name path=lower, blue, thick, forget plot, domain=-1:0, samples=200] {2*x^2 - 1};
\addplot[name path=lower, green, thick, forget plot, domain=0:1, samples=200] {2*x^2 - 1};
\addplot[name path=lower,green!25!blue!75, thick, forget plot, domain=-0.5:0, samples=200] {2*(x/(2*0.25-1))^2 - 1};

% Parametric curves

\addplot[
    only marks,
    mark=square*,
    mark size=2pt,
    yellow,
] coordinates {
    ({-0.353},{2*0.353^2 - 1})
    ({-0.1768},{2*(0.1768/(2*0.25 - 1))^2 - 1})
    ({0.353},{2*0.353^2 - 1})

};

\addplot[
    only marks,
    mark=square*,
    mark size=2pt,
    red
] coordinates {
    ({-0.935},{2*0.935^2 - 1})
    ({-0.4677},{2*(0.4677/(2*0.25 - 1))^2 - 1})
    ({0.935},{2*0.935^2 - 1})
};

% Curve 1
\addplot[
	mesh,	
	ultra thick,
    domain=0:1,
    variable=t,
    samples=200, % required but unused
    shader=interp,
    point meta=t,
    ]({-(t*0.582+0.353)},{2*(t*0.582+0.353)^2 - 1});

\addplot[
	mesh,	
    ultra thick,
    domain=0:1,
    samples=200,
    variable=t,
    shader=interp,
    point meta=t,
] ({(t*0.582+0.353)},{2*(t*0.582+0.353)^2 - 1});

\addplot[
	mesh,
    ultra thick,
    domain=0:1,
    samples=200,
    variable=t,
    shader=interp,
    point meta=t,
    ] ({-(0.2909*t+0.1768)},{2*((0.2909*t+0.1768)/(2*0.25 - 1))^2 - 1});

%% P1 and P2 markers
%\addplot[only marks, mark=square*, mark size=2.5pt, color=yellow] coordinates {(0,-1)};
%\addplot[only marks, mark=*, mark size=2.5pt, color=red] coordinates {(-0.75,0.75) (0.75,0.75)};

% Dashed lines
\addplot[dashed, red] coordinates {(-0.935,0.75) (0.935,0.75)};
%\addplot[dashed, yellow] coordinates {(0,-1) (-0.75,-1)};
\addplot[dashed, yellow] coordinates {(-0.353,-0.75) (0.353,-0.75)};

% Text labels
\node[blue] at (axis cs:-0.8,1.075) {\small $\theta_l: 45\mbox{-}90$};
\node[green] at (axis cs:1,1.075) {\small $\theta_r: 45\mbox{-}0$};
\node[green!25!blue!75] at (axis cs:-0.215,1.075) {\small $R_l = 0.75$};
\node[green!25!blue!75] at (axis cs:-0.215,0.925) {\small $R_r = 0.25$};
\node[yellow] at (axis cs:-0.1,-0.65) {$P_1$};
\node[red] at (axis cs:-0.35,0.65) {$P_2$};
\end{axis}
\end{tikzpicture}}
    	\caption{Example interpolation between two laminates for volume fractions \( R_l = 0.75 \), \( R_r = 0.25 \). Design points: \( P_1(V_1, V_3) = (-0.1768, -0.75) \), \( P_2(V_1, V_3) = (-0.4677, 0.75) \).}
    \label{fig:Miki_Lamination_Diagram_Interpolation_1}
   \end{subfigure}
   \caption{Examples of Interpolation Curves by using the Lamination Parameter Interpolation Method. Adapted from~\cite{serhat_lamination_2019}.}
   \label{fig:Miki_Lamination_Diagram_Interpolation}
\end{figure}
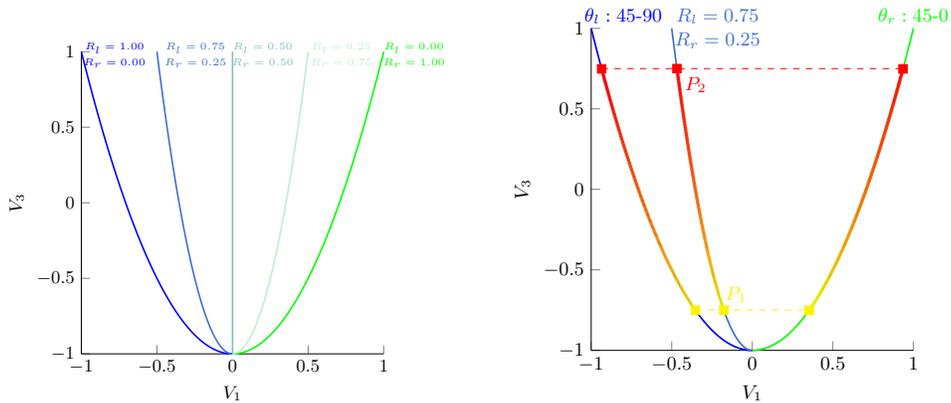

%\paragraph{Lamination Parameter Distribution Generation.}  
% In the first step, an optimal distribution of LPs is determined while maintaining feasibility. The variation of LPs is restricted to lie on prescribed curves in Miki’s diagram, which represent valid stacking sequences as functions of the volumetric ratio of ply angles. Specifically, when considering at most two different fiber angle orientations, the relationship between the LPs and the volumetric ratio \( R_r \in [0, 1] \) is given by:
When using at most two unique sets of fiber orientations, the LP design space is characterized by curves in the \( (V_1,V_3) \) plane, each corresponding to a specific volumetric ratio \( R_r \in [0, 1] \), which denotes the volume fraction associated with ply angles from the right boundary of Miki’s diagram (i.e., the fraction of plies oriented between \( 45^\circ \) and \( 0^\circ \)). The corresponding left-side volume fraction is \( R_l = 1 - R_r \). A set of such curves is illustrated in Figure~\ref{fig:Miki_Lamination_Diagram_Interpolation_2}, showing the feasible design space for various \( R_r \).
The relation between \( V_1, V_3, \text{ and } R_r\) is given by:
\begin{equation}
V_1 = (2 R_r - 1) \sqrt{\frac{V_3 + 1}{2}},
\label{eq:Relation_to_volumetric_ratio_and_lps}
\end{equation}
This constraint ensures that the resulting stacking sequences are physically realizable. 
% where \( R_r \) denotes the volume fraction associated with ply angles from the right boundary of Miki’s diagram (i.e., \( 45^\circ \) to \( 0^\circ \)), and the corresponding left-side volume fraction is \( R_l = 1 - R_r \). A set of such curves is illustrated in Figure~\ref{fig:Miki_Lamination_Diagram_Interpolation_2}, showing the feasible design space for various \( R_r \).

To allow smooth spatial variation across the structure, we interpolate LPs between selected design points
\(P_1\) and \(P_2\) on these curves using a distance-weighted scheme. The interpolated point satisfies an arc-length proportionality condition (\(\widehat{P_e P_1}/{\widehat{P_e P_2}} = {d_1}/{d_2}\)), ensuring geometric consistency and manufacturability (see Figure~\ref{fig:Miki_Lamination_Diagram_Interpolation_1}).

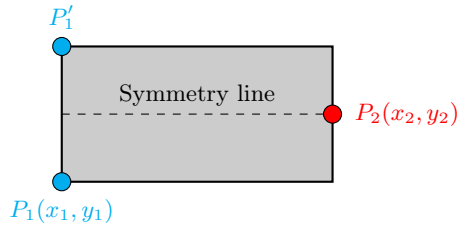
\begin{figure}[htbp]
    \centering
    \begin{tikzpicture}[scale=0.6]

% Draw rectangle
\fill[gray!40] (0,0) rectangle (6,3);
\draw[thick] (0,0) rectangle (6,3);

% Draw dashed line
\draw[dashed] (0,1.5) -- (6,1.5) node[midway, above] {Symmetry line};;

% Draw P1 point (bottom-left)
\fill[fill=cyan, draw=black] (0,0) circle (0.2);
\node[below, text=cyan] at (0,-0.2) {$P_1(x_1,y_1)$};

% Draw P1 point (bottom-left)
\fill[fill=cyan, draw=black] (0,3) circle (0.2);
\node[above, text=cyan] at (0,3.2) {$P_1'$};

% Draw P2 point (center)
\fill[fill=red, draw=black] (6,1.5) circle (0.2);
\node[right, text=red] at (6.3,1.5) {$P_2(x_2,y_2)$};

\end{tikzpicture}
    \caption{Master nodes used for the interpolation of LPs.}
    \label{fig:master_nodes}
\end{figure}
\section{Optimization Problem}
\label{sec:opti_problem}
\subsubsection{Design Variables.}
\label{subsubsec:design_variables}
We define the design vector as
$ \mathbf{x} = \left\{ \mathbf{x}_{\text{MMC}}, \mathbf{x}_{\text{LP}} \right \}   \in \mathbb{R}^n $ where
$  \mathbf{x}_{\text{MMC}} \in \mathbb{R}^{n_1} $ are parameters that define the distribution of the material through a collection of MMCs, whereas \( \mathbf{x}_{\text{LP}} \in \mathbb{R}^{n_2} \) are LPs encoding in-plane stiffness properties, namely a vector comprising of 
(\( R_{r}, V_{3,1}, V_{3,2} \)), with $R_r$ as the volumetric ratio and $V_{3,1}$ and $V_{3,2}$ correspond to the $V_3$ values fixed at the master nodes shown in Figure \ref{fig:master_nodes}.
In this study, we consider three MMCs with symmetry assumptions, leading to $n_{1}=15$. With the three additional LPs, the total number of design variables is \(n=18 \).

\subsubsection{Objective Function.}
\label{subsubsec:objective_function}
The goal is to minimize structural compliance:
\begin{equation}
\min_{\mathbf{x}} \; C(\mathbf{x}) = \mathbf{f}^\top \mathbf{u}_{\text{global}} = \sum_{k}^{n_{\text{elements}}} {\mathbf{u}}_{(k)}^{\top} \mathbf{K}_{(k)} {\mathbf{u}}_{(k)}.
\label{eq:compliance_definition}
\end{equation}
where ${\mathbf{u}}_{(k)}$ and $\mathbf{K}_{(k)}$ denote the displacement vector and stiffness matrix of element \(k\), respectively.
%
% With some abuse of notation, ${\mathbf{u}}_{(k)}$ and $\mathbf{K}_{(k)}$ correspond to the displacement vector and stiffness matrix of the $k$-th element, respectively. However, this is practical since this shows that the compliance $C(\mathbf{x}) > 0$ given that each stiffness matrix $\mathbf{K}_{(k)}$ is positive definite.
%
%
\subsubsection{Constraints.}
\label{subsubsec:constraints}
Optimization is subject to the following constraints:
\begin{enumerate}
%     \item \textsc{Static Condition:}
%     \begin{equation}
% \mathbf{K}_{\text{global}}(\mathbf{x}) \mathbf{u}_{\text{global}} = \mathbf{f} \; ,
% \end{equation}
% where \( \mathbf{K}_{\text{global}}(\mathbf{x}) \) is the global stiffness matrix dependent on geometry and material anisotropy, $\mathbf{u}_{\text{global}}$ the global displacement vector and $\mathbf{f}$ the external load vector.
    \item \textsc{Volume Constraint:}
\begin{equation}
	V(\mathbf{x}) -  V_{\text{max}} \leq 0
	\label{eq:volume_constraint}
\end{equation}
Here, \( V(\mathbf{x}) \) is the volume defined by the MMC geometry. For this work, we just apply a volume constraint $V_{\text{max}}$ of 50\% of the design domain. The function \( V(\mathbf{x}) \) is computed by counting the discrete elements classified as part of $\Omega$.

    \item \textsc{Connectivity Constraint:}
    \begin{equation}
        \Psi(\mathbf{x}) = \psi_1(\mathbf{x}) +\psi_2(\mathbf{x}) + \psi_3(\mathbf{x})  \leq 0
    \label{eq:connectivity_constraint}
    \end{equation}
    
    Here, \( \Psi \) serves as a penalty function quantifying the extent to which the structure exhibits disconnection. We adopt three different independent measures shown in Figure \ref{fig:disconnected_figures}. As distance-based quantities, all individual terms satisfy $\psi_i \geq 0$.    
For the numerical evaluation of these measures, we employed the \texttt{Python} package \texttt{shapely} \cite{shapely2024}. 
%Using this library, the boundary $\partial \Omega$ of the global LSF $\Phi(\mathbf{q})$ was discretized, enabling the computation of the minimum distances from the load application point ($\psi_1$) and the left boundary of the domain ($\psi_2$) to $\partial \Omega$. The third measure, $\psi_3$, is nonzero only when $\partial \Omega$ is not a single closed contour but comprises multiple disjoint components. In such cases, $\psi_3$ is defined as the sum of the minimum distances from each disjoint component to its nearest neighbour, thereby quantifying the degree of discontinuity in the boundary.

% For the numerical evaluation of these measures, we employed the \texttt{Python} package \texttt{shapely} \cite{shapely2024}. Using this library, the boundary $\partial \Omega$ of the global LSF $\Phi(\mathbf{q})$ was discretized, enabling the computation of the minimum distances from the load application point ($\psi_1$) and the left boundary of the domain ($\psi_2$) to $\partial \Omega$. The third measure, $\psi_3$, is nonzero only when $\partial \Omega$ is not a single closed contour but comprises multiple disjoint components. In such cases, $\psi_3$ is defined as the sum of the minimum distances from each disjoint component to its nearest neighbour, thereby quantifying the degree of discontinuity in the boundary.

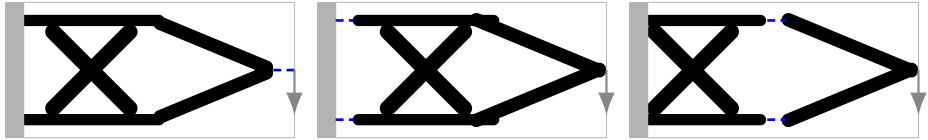
\begin{figure}[htbp]
\centering
\tikzset{every picture/.style={scale=0.6, transform shape}}

\begin{subfigure}[t]{0.32\textwidth}
    \centering
    \raisebox{2.5mm}{\begin{tikzpicture}
%
% Correct capsule slot macro: rectangle + 2 semicircles
\newcommand{\slot}[5]{%
  \pgfmathsetmacro{\L}{#4}
  \pgfmathsetmacro{\W}{#5}
  \pgfmathsetmacro{\R}{0.5 * \W}
  \begin{scope}[shift={(#1,#2)}, rotate=#3]
    \path[fill=black]
      ({-\L/2 + \R}, \R)
        arc[start angle=90, end angle=270, radius=\R] --
      ({\L/2 - \R}, -\R)
        arc[start angle=90, end angle=270, radius=-\R] --
      cycle;
  \end{scope}
}
%
% Fixed slot - now both ends are correct
\slot{4.2}{0.95}{22.5}{2.85}{0.3}
\slot{4.2}{2.05}{337.5}{2.85}{0.3}

\slot{1.5}{0.4}{0}{3.25}{0.25}
\slot{1.5}{2.6}{0}{3.25}{0.25}

\slot{1.5}{1.5}{45}{2.75}{0.35}
\slot{1.5}{1.5}{315}{2.75}{0.35}

% Test domain
\draw[thin, gray!55] (0,0) rectangle (6,3);
\fill[gray!60] (0,0) rectangle (-0.4,3);
\draw[blue, densely dashed, line width=1pt] (6.0,1.5) -- ++(-0.5,0);
%\draw[blue, dashed, line width=1pt] (0.0,2.6) -- ++(0.4,0);
\draw[gray!95, thick, -{Latex[length=3mm]}] (6.0,1.5) -- ++(0,-1);
\end{tikzpicture}}
    \caption{Connection to the load, \(\psi_1\).}
    \label{fig:first}
\end{subfigure}
\hfill
\begin{subfigure}[t]{0.32\textwidth}
    \centering
    \raisebox{2.5mm}{\begin{tikzpicture}
%
% Correct capsule slot macro: rectangle + 2 semicircles
\newcommand{\slot}[5]{%
  \pgfmathsetmacro{\L}{#4}
  \pgfmathsetmacro{\W}{#5}
  \pgfmathsetmacro{\R}{0.5 * \W}
  \begin{scope}[shift={(#1,#2)}, rotate=#3]
    \path[fill=black]
      ({-\L/2 + \R}, \R)
        arc[start angle=90, end angle=270, radius=\R] --
      ({\L/2 - \R}, -\R)
        arc[start angle=90, end angle=270, radius=-\R] --
      cycle;
  \end{scope}
}
%
% Fixed slot - now both ends are correct
\slot{4.48}{0.95}{22.5}{3.25}{0.3}
\slot{4.48}{2.05}{337.5}{3.25}{0.3}
\slot{2}{0.4}{0}{3.25}{0.25}
\slot{2}{2.6}{0}{3.25}{0.25}
\slot{2}{1.5}{45}{2.75}{0.35}
\slot{2}{1.5}{315}{2.75}{0.35}
%
% Test domain
\draw[thin, gray!55] (0,0) rectangle (6,3);
\fill[gray!60] (0,0) rectangle (-0.4,3);
\draw[blue, densely dashed, line width=1pt] (0.0,0.4) -- ++(0.4,0);
\draw[blue, densely dashed, line width=1pt] (0.0,2.6) -- ++(0.4,0);
\draw[gray!95, thick, -{Latex[length=3mm]}] (6.0,1.5) -- ++(0,-1);
\end{tikzpicture}}
    \caption{Connection to the support, \(\psi_2\).}
    \label{fig:second}
\end{subfigure}
\hfill
\begin{subfigure}[t]{0.32\textwidth}
    \centering
    \raisebox{2.5mm}{\begin{tikzpicture}

% Correct capsule slot macro: rectangle + 2 semicircles
\newcommand{\slot}[5]{%
  \pgfmathsetmacro{\L}{#4}
  \pgfmathsetmacro{\W}{#5}
  \pgfmathsetmacro{\R}{0.5 * \W}
  \begin{scope}[shift={(#1,#2)}, rotate=#3]
    \path[fill=black]
      ({-\L/2 + \R}, \R)
        arc[start angle=90, end angle=270, radius=\R] --
      ({\L/2 - \R}, -\R)
        arc[start angle=90, end angle=270, radius=-\R] --
      cycle;
  \end{scope}
}

% Fixed slot - now both ends are correct
\slot{4.48}{0.95}{22.5}{3.25}{0.3}
\slot{4.48}{2.05}{337.5}{3.25}{0.3}

\slot{1.25}{0.4}{0}{2.75}{0.25}
\slot{1.25}{2.6}{0}{2.75}{0.25}

\slot{1.}{1.5}{45}{2.75}{0.35}
\slot{1.}{1.5}{315}{2.75}{0.35}

% Test domain
\draw[thin, gray!55] (0,0) rectangle (6,3);
\fill[gray!60] (0,0) rectangle (-0.4,3);
\draw[blue, densely dashed, line width=1pt] (2.65,0.4) -- ++(0.4,0);
\draw[blue, densely dashed, line width=1pt] (2.65,2.6) -- ++(0.4,0);
\draw[gray!95, thick, -{Latex[length=3mm]}] (6.0,1.5) -- ++(0,-1);

\end{tikzpicture}} 
    \caption{Connection of the structure itself, \(\psi_3\).}
    \label{fig:third}
\end{subfigure}

\caption{Exemplary violations of the connectivity constraint in the cantilever beam test case. The gaps preventing the design from satisfying the connectivity criterion are indicated by the \textcolor{blue}{blue} dashed lines. Adapted from \cite{raponi_kriging-assisted_2019}.}
\label{fig:disconnected_figures}
\end{figure}

%     \item \textsc{Lamination Feasibility:}
% \begin{equation}
% 	M_j(\mathbf{x}_{\text{LP}}) \in \mathcal{L}
% 	\label{eq:lamination_feasibility}
% \end{equation}

% Let $M_j$ denote the mapping from the design variables $(\mathbf{x}_{\text{LP}})$ to the LPs associated with each finite element discretizing the domain $D$. The set $\mathcal{L}$ represents the admissible region of LPs defined by Miki’s diagram. While Equation~\eqref{eq:lamination_feasibility} expresses the general requirement that the LPs must remain within this feasible region, the LPIM procedure described in Section~\ref{subsubsec:LPIM} ensures that this condition is inherently satisfied by construction, as the parameterization restricts the design space to $\mathcal{L}$ throughout the optimization process.
\end{enumerate}

%\subsection{Complete Optimization Problem}
%
%The overall optimization problem is thus:
%\[
%\begin{aligned}
%\min_{\mathbf{x}} \quad & C(\mathbf{x}) = \mathbf{f}^\top \mathbf{K}^{-1}(\mathbf{x}) \mathbf{f} \\
%\text{s.t.} \quad & V(\mathbf{x}) \leq V_{\text{max}} \\
%& \Phi(\mathbf{x}) \geq 0 \\
%& \mathbf{x}_{\text{lp}} \in \mathcal{L}
%\end{aligned}
%\]

\subsubsection{Modified Objective Function. }
\label{subsubsec:modified_objective_function}
To incorporate the constraints, we adopt a modified compliance objective given by:
\begin{equation}
\tilde{C}(\mathbf{x}) =
\begin{cases}
C(\mathbf{x}) + \gamma_1 \max\!\bigl(V(\mathbf{x})-V_{\max},0\bigr),
& \Psi(\mathbf{x}) \le 0, \\[4pt]
\gamma_1 \max\!\bigl(V(\mathbf{x})-V_{\max},0\bigr) + \gamma_2 \Psi(\mathbf{x}),
& \Psi(\mathbf{x}) > 0,
\end{cases}
\label{eq:modified_compliance}
\end{equation}
with $\gamma_1=0.02$ and $\gamma_2=200$.

This formulation serves two primary purposes. First, it improves computational efficiency by avoiding evaluations of the compliance \( C(\mathbf{x}) \) when the structure is disconnected. In such cases, the finite element solver may yield numerically unstable or uninformative results, thereby introducing randomness into the search process. By penalizing based on the infeasibility measure \( \Psi(\mathbf{x}) \), the optimizer can more effectively steer the search toward feasible and connected designs. 

Second, this modification transforms the constrained optimization problem into an unconstrained one, which is particularly advantageous for population-based optimization algorithms. These algorithms typically lack inherent mechanisms to handle hard constraints, and penalty formulations like the one in Equation~\eqref{eq:modified_compliance} offer a practical workaround. However, this transformation introduces potential drawbacks for model-based optimizers, such as those used in BO. The resulting objective function may exhibit discontinuities and non-smooth behavior, complicating surrogate model training and acquisition function optimization. However, to maintain generality and facilitate performance comparisons across different classes of optimizers, we ultimately adopt the aforementioned formulation.
% While several constraint-handling strategies have been proposed in the BO literature~\cite{gardner2014,gramacy_modeling_2016,letham_constrained_2019,picheny_bayesian_2016}, the majority are limited to low-dimensional problems or standard (vanilla) BO formulations. To the best of our knowledge, only the approaches introduced by Ascia et al.~\cite{ascia_feasibility-driven_2025} and Eriksson et al.~\cite{eriksson_scalable_2021} offer scalable methods capable of addressing constraints in high-dimensional design spaces, thereby underscoring a persistent gap in current BO methodologies. 
% To maintain generality and avoid restricting our methodology to a narrow class of optimizers, we ultimately adopt the modified compliance formulation defined in~\eqref{eq:modified_compliance}.

\section{Experimental Setup}
\label{sec:experimental_setup}
%
% We aim at coupling a set of black-box optimizers with the problem stated in Section \ref{sec:Problem Formulation} whilst sampling the modified objective function described in Equation \eqref{eq:modified_compliance}.

We evaluate two distinct strategies (concurrent and sequential) for solving the problem under a fixed total function evaluation budget $B$ (with $B=1000$ for the experiments presented here). For both strategies, we used the following material properties: a longitudinal Young's modulus $E_{1}=25$, a transverse Young's modulus $E_{2}=1$, an in-plane shear modulus $G_{12}=0.5$, and a Poisson's ratio $\nu_{12}=0.25$. These values are consistent with the proportions presented in \cite{serhat_lamination_2019}.
% %
% \begin{table}[htbp!]
% %\begin{wraptable}{r}{3.5cm}
% 	\centering
% 	\caption{Material properties of the laminae}
% 	\begin{tabular}{|l|l|}
% 	\hline
% 	\textbf{Property}   & \textbf{Value} \\ \hline \hline
% 	$E_{1}$   & 25 \\ \hline
% 	$E_{2}$   & 1  \\ \hline
% 	$G_{12}$   & 0.5 \\ \hline
% 	$\nu_{12}$ & 0.25 \\ \hline
% 	\end{tabular}
% 	\label{tab:Material_Properties}
% %\end{wraptable}
% \end{table}
%

\subsubsection{Concurrent Approach.}
\label{subsubsec:concurrent_approach}
The entire set of $n=18$ design variables (15 topology variables and 3 LPs) is optimized simultaneously. 
Each algorithm operates directly in this joint 18-dimensional search space and is given the full budget.

\subsubsection{Sequential Approach.}
\label{subsubsec:sequential_approach}
The optimization is decomposed into two stages:
\begin{enumerate}
\item Optimize the $n_1=15$ topology variables using a partial budget. We computed the partial budget $B_1$ as a weighed proportion of the dimensions of $\mathbf{x}_{\text{MMC}}$ and $\mathbf{x}_{\text{LP}}$ as:
\begin{equation}
B_1 = B \frac{n_1}{n} = B \frac{n_1}{n_1+n_2}.
\label{eq:weighed_proportion}
\end{equation}
For this part, we fixed the vector $\mathbf{x}_{\text{LP}}$ to \( \left(R_r=0.5,V_{3,1}=0,V_{3,2}=0 \right) \) such that the material is quasi-isotropic. This particular approach was also used by Peeters et al.~\cite{peeters_combining_2015}.
\item Fix the best topology obtained from stage 1 and optimize the 3 LP variables using the remaining budget $B_2=B-B_1$. 
\end{enumerate}

\subsection{Algorithms}
\label{subsec:algorithms}
We compare the following black-box optimization algorithms for both the concurrent and sequential strategies: 1) CMA-ES; 2) BAxUS; 3) HEBO; 4) vanilla BO; 5) TuRBO-1; 6) TuRBO-m; 7) Differential Evolution (DE). 
%Since each optimization strategy has different dimensions, we present some technicalities of each of the algorithms as sections.

\subsubsection{CMA-ES.} 
\label{subsubsec:CMA-ES}
In this algorithm, the offspring population size $\lambda$ is defined as:
\(\lambda = 4 + \lfloor 3 \ln d \rfloor\), 
where $d$ is the problem dimension and $\lfloor \cdot \rfloor$ the floor operator. This value adapts to the dimensionality associated with the concurrent and sequential optimization problem setups. The parent population size is set to $\mu = \lambda / 2$, following the default configuration of the \texttt{PyCMA} library \cite{hansen_cma_2006}. Additionally, the active CMA-ES variant \cite{jastrebski_improving_2006} is used due to its high performance on multimodal optimization problems.
\subsubsection{DE.}
\label{subusubsec:DE}
For this algorithm, we set the population size to \(10 d\). This choice ensures adequate diversity to effectively explore the search space, particularly in higher dimensions. The binomial crossover strategy ("bin") was used, along with the default parameter settings from the \texttt{scipy.optimize.differential\_evolution} implementation: a crossover probability \(CR = 0.7\) and a differential weight \(F = 0.8\). 
\subsubsection{BO.}
\label{subusubsec:BO}
For BO methods, the initial sample set size is set to $3d$. To ensure consistent and serial evaluation across all methods, no parallelism was considered. All optimizers employed a Gaussian Process surrogate with a Matérn $5/2$ kernel; Automatic Relevance Determination (ARD) was applied to adapt lengthscales individually for each input dimension, except in the case of BAxUS.

Table~\ref{tab:bayesian_descriptors} lists the algorithms, their corresponding acquisition functions, and the Python libraries from which their implementations are sourced.
\begin{table}[htbp]
\centering
\caption{List of BO methods showing acquisition function and implementation framework.}
\begin{tabular}{|l|p{18em}|l|}
\hline
\textbf{Optimizer}  & \textbf{Acquisition Function}                                                                 & \textbf{Library} \\ \hline
\textbf{TuRBO-1}    & Thompson Sampling \cite{thompson_likelihood_1933}                                                                         & BoTorch          \\
\textbf{TuRBO-m}    & Thompson Sampling \cite{thompson_likelihood_1933}                                                                         (with 3 multi-trust-region variant of TuRBO-1)                                      & BoTorch          \\
\textbf{HEBO}       & Multi-objective Acquisition Ensemble (MACE) \cite{cowen-rivers_hebo_2022}                     & HEBO             \\
\textbf{BAxUS}      & Thompson Sampling \cite{thompson_likelihood_1933}                                                                          & BoTorch          \\
\textbf{vanilla BO} & Log Expected Improvement (logEI) \cite{ament_unexpected_2025}                                  & BoTorch          \\ \hline
\end{tabular}
\label{tab:bayesian_descriptors}
\end{table}
\section{Results}
\label{sec:results}
\subsection{Convergence and Global Results}
\label{subsec:global_results}
\begin{figure}[htbp]
    \centering
    % Placeholder box (optional)
	%\includegraphics[width=\textwidth]{./Figures/Sequential_v_Concurrent.pdf}
    \begin{tikzpicture}

% Main plot
\begin{axis}[
    name=mainplot,
    xlabel={Function Evaluations},
    ylabel={Modified compliance [best-so-far]},
    ymin=5e-2, ymax=1.3e6, % Adjust this to your actual range
    xmax=1000,
    xmin=0,
    grid=both,
    major grid style={line width=0.4pt, color=gray!20},
    minor grid style={line width=0.2pt, color=gray!10},
    minor x tick num=1,
    minor y tick num=1,
    ymode=log,
    ytick pos=left,
    xtick pos=bottom,
    width=0.65\linewidth,
      legend style={
        at={(0.5,-0.2)},
        anchor=north,
        legend columns=8,
        draw=none,
        fill=none,
        font=\tiny,
    },
]

% Add vertical line
\addplot[
    domain=5e-2:2e6,
    samples=2,
    ultra thick,
    black,
    dotted,
    forget plot
] ({833}, x);

% Concurrent (solid)
\addplot[no marks, thick, blue] table [x=evaluations, y=mean, col sep=comma] {./Tikzplots/convergence_data/CMA_ES_Concurrent_orthotropic.csv};
% \addlegendentry{}
% Sequential (dashed)
\addplot[no marks, thick, blue, dashed] table [x=evaluations, y=mean, col sep=comma] {./Tikzplots/convergence_data/CMA_ES_Sequential_orthotropic.csv};
% \addlegendentry{CMA-ES}
% Concurrent (solid)
\addplot[no marks, thick, red] table [x=evaluations, y=mean, col sep=comma] {./Tikzplots/convergence_data/DE_Concurrent_orthotropic.csv};
% \addlegendentry{}
% Sequential (dashed)
\addplot[no marks, thick, red, dashed] table [x=evaluations, y=mean, col sep=comma] {./Tikzplots/convergence_data/DE_Sequential_orthotropic.csv};
% \addlegendentry{DE}
% Concurrent (solid)
\addplot[no marks, thick, magenta] table [x=evaluations, y=mean, col sep=comma] {./Tikzplots/convergence_data/HEBO_Concurrent_orthotropic.csv};
% \addlegendentry{}
% Sequential (dashed)
\addplot[no marks, thick, magenta, dashed] table [x=evaluations, y=mean, col sep=comma] {./Tikzplots/convergence_data/HEBO_Sequential_orthotropic.csv};
% \addlegendentry{HEBO}
% Concurrent (solid)
\addplot[no marks, thick, yellow] table [x=evaluations, y=mean, col sep=comma] {./Tikzplots/convergence_data/turbo_1_Concurrent_orthotropic.csv};
% \addlegendentry{}
% Sequential (dashed)
\addplot[no marks, thick, yellow, dashed] table [x=evaluations, y=mean, col sep=comma] {./Tikzplots/convergence_data/turbo_1_Sequential_orthotropic.csv};
% \addlegendentry{TuRBO-1}
% Concurrent (solid)
\addplot[no marks, thick, green] table [x=evaluations, y=mean, col sep=comma] {./Tikzplots/convergence_data/turbo_m_Concurrent_orthotropic.csv};
% \addlegendentry{}
% Sequential (dashed)
\addplot[no marks, thick, green, dashed] table [x=evaluations, y=mean, col sep=comma] {./Tikzplots/convergence_data/turbo_m_Sequential_orthotropic.csv};
% \addlegendentry{TuRBO-m}
% Concurrent (solid)
\addplot[no marks, thick, violet] table [x=evaluations, y=mean, col sep=comma] {./Tikzplots/convergence_data/Vanilla_BO_Concurrent_orthotropic.csv};
% \addlegendentry{}
% Sequential (dashed)
\addplot[no marks, thick, violet, dashed] table [x=evaluations, y=mean, col sep=comma] {./Tikzplots/convergence_data/Vanilla_BO_Sequential_orthotropic.csv};
% \addlegendentry{Vanilla BO}
% Concurrent (solid)
\addplot[no marks, thick, brown] table [x=evaluations, y=mean, col sep=comma] {./Tikzplots/convergence_data/BAxUS_Concurrent_orthotropic.csv};
% \addlegendentry{}
% Sequential (dashed)
\addplot[no marks, thick, brown, dashed] table [x=evaluations, y=mean, col sep=comma] {./Tikzplots/convergence_data/BAxUS_Sequential_orthotropic.csv};
% \addlegendentry{BAxUS}
%
\end{axis}

% Manual legend
\node[anchor=north west] at ([xshift=0.2cm,yshift=-1cm]mainplot.north east) {
\scalebox{0.52}{
\begin{tikzpicture}
\matrix[anchor=west,ampersand replacement=\&, row sep=3pt, column sep=20pt] {
    \node[draw=none] {\textbf{Algorithm}}; \& \node[draw=none, xshift=-5pt] {\textbf{Concurrent}}; \& \node[draw=none, xshift=-5pt] {\textbf{Sequential}}; \\
    \node[draw=none] {CMA-ES};        \& \draw[blue, thick] (0,0) -- +(1,0); \& \draw[blue, thick, dashed] (0,0) -- +(1,0); \\
    \node[draw=none] {DE};            \& \draw[red, thick] (0,0) -- +(1,0); \& \draw[red, thick, dashed] (0,0) -- +(1,0); \\
    \node[draw=none] {HEBO};          \& \draw[magenta, thick] (0,0) -- +(1,0); \& \draw[magenta, thick, dashed] (0,0) -- +(1,0); \\
    \node[draw=none] {TuRBO-1};       \& \draw[yellow, thick] (0,0) -- +(1,0); \& \draw[yellow, thick, dashed] (0,0) -- +(1,0); \\
    \node[draw=none] {TuRBO-m};       \& \draw[green, thick] (0,0) -- +(1,0); \& \draw[green, thick, dashed] (0,0) -- +(1,0); \\
    \node[draw=none] {vanilla BO};    \& \draw[violet, thick] (0,0) -- +(1,0); \& \draw[violet, thick, dashed] (0,0) -- +(1,0); \\
    \node[draw=none] {BAxUS};         \& \draw[brown, thick] (0,0) -- +(1,0); \& \draw[brown, thick, dashed] (0,0) -- +(1,0); \\
};

\end{tikzpicture}
}};
\end{tikzpicture}
    \caption{Convergence plot of the algorithms under concurrent (solid lines) and sequential (dashed lines) approaches, averaged over 20 runs. The black dotted line marks the transition between the two stages of the sequential strategy.}
    \label{fig:Convergence_Plots_global}
\end{figure}
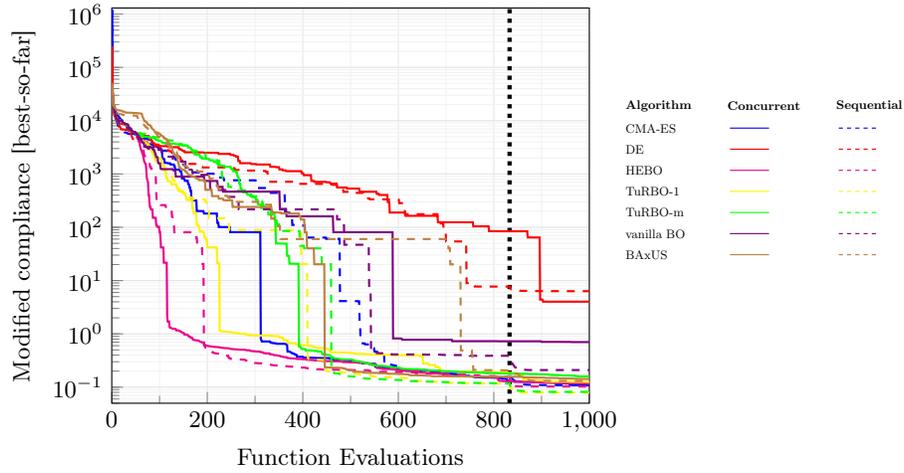

Figure~\ref{fig:Convergence_Plots_global} presents the convergence performance of several optimizers under both sequential and concurrent evaluation regimes. The vertical axis shows the best-so-far objective value (lower is better), while the horizontal axis indicates the number of function evaluations.

In general, sequential variants consistently achieve better final solutions compared to their concurrent counterparts, except for DE. Notably, the concurrent strategy demonstrates better convergence with fewer evaluations. This result is not surprising as the quasi-isotropic material definition is suboptimal for most topologies. In this context, the concurrent strategy benefits from its ability to adjust the fiber orientations, leading to improved compliance. 
The HEBO optimizer achieves the best performance for small budgets, specifically for the first 400 evaluations. Within this range, both the sequential and concurrent variants of HEBO outperformed the other methods, indicating that HEBO quickly guides the search toward the feasible region of the design space. However, its relative performance declines as the number of evaluations increases. In particular, the sequential versions of TuRBO-1 and TuRBO-m show the best average performance for larger budgets.
Furthermore, the sequential convergence curves suggest that the allocated budget for stage 2 may exceed what is necessary. Most of the improvement occurs early, with only marginal refinement observed toward the end. This behavior reflects the convex nature of the optimization problem when formulated in terms of LPs alone, making it easier for the optimizers to find a near-optimal configuration.
\begin{figure}[htbp]
    \centering
    % Placeholder box (optional)
%	\includegraphics[width=\linewidth]{./Figures/box_plot_algo_full_budget_wo_mean.pdf}
\def\svgwidth{0.8\linewidth}
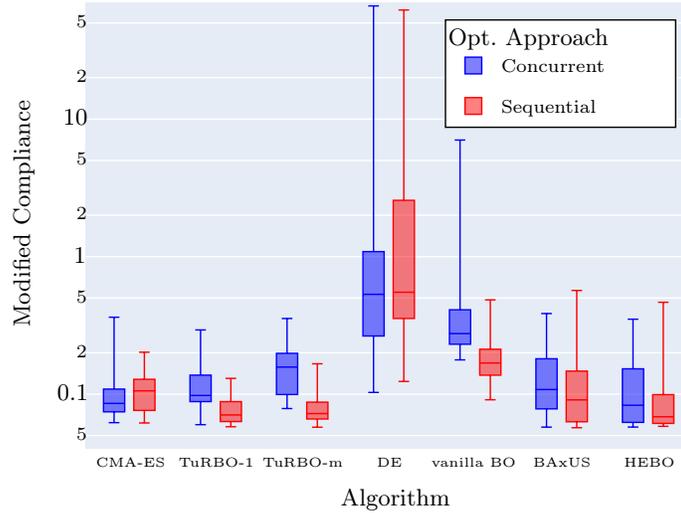
    %%{./Figures/box_plot_algo_full_budget.pdf}
    \caption{Boxplots of modified compliance values for each optimization algorithm under sequential and concurrent strategies after the full evaluation budget. Lower values indicate better performance. }
    \label{fig:box_plots_compliance}
\end{figure}
To complement the observations from Figure~\ref{fig:Convergence_Plots_global}, Figure~\ref{fig:box_plots_compliance} presents the distribution of modified compliance values for the best-performing structures, grouped by optimization strategy and algorithm. The plot clearly demonstrates that sequential strategies generally achieve lower median compliance and exhibit reduced variability compared to their concurrent counterparts, particularly for TuRBO-1, TuRBO-m and vanilla BO. To statistically assess whether the medians obtained using the concurrent and sequential approaches differ for each algorithm, a Wilcoxon signed-rank test was conducted. The results, summarized in Table~\ref{tab:wilcoxon}, show that the null hypothesis stating that the medians of both approaches are equal was rejected at the 5\% significance level.

\begin{table}[htbp]
\centering
\caption{Wilcoxon signed-rank test results comparing concurrent and sequential strategies in terms of median optimal compliance. Statistically significant results ($p < 0.05$) are highlighted in bold.}
\label{tab:wilcoxon}
\begin{tabular}{|l |  S[scientific-notation=true, round-mode=places, round-precision=5] | S[scientific-notation=true, round-mode=places, round-precision=5] | c |}
\hline
\textbf{Algorithm} & \textbf{Concurrent Median} & \textbf{Sequential Median} & \textit{p}-\textbf{value} \\
\hline
CMA-ES     & 8.56575e-2 & 1.05749e-1 & 0.7012 \\
TuRBO-1    & 9.78679e-2 & 7.05895e-2 & \textbf{0.0107} \\
TuRBO-m    & 1.57328e-1 & 7.22293e-2 & \textbf{0.0003} \\
DE         & 5.31994e-1 & 5.50749e-1 & 0.2774 \\
vanilla BO & 2.75778e-1 & 1.68549e-1 & \textbf{0.0107} \\
BAxUS      & 1.08034e-1 & 9.08690e-2 & 0.3683 \\
HEBO       & 8.30844e-2 & 6.84599e-2 & 0.1231 \\
\hline
\end{tabular}
\end{table}

% \begin{table}[hb!]
% \centering
% \setlength{\tabcolsep}{6pt} % tighter spacing
% \renewcommand{\arraystretch}{1.3} % row spacing
% \caption{Wilcoxon Signed-Rank Test: Sequential vs Concurrent (\textit{p}-values)}
% \label{tab:wilcoxon}
% \begin{tabularx}{\textwidth}{l*{6}{>{\centering\arraybackslash}p{1.5cm}}}
% \hline
% CMA-ES & TuRBO-1 & TuRBO-m & DE & Vanilla-BO & BAxUS & HEBO \\ \hline
% 0.7012 & 0.0107 & 0.0003 & 0.2774 & 0.0107 & 0.3683 & 0.1231 \\ \hline
% \end{tabularx}
% \end{table}

The marked performance gap between the sequential and concurrent variants of TuRBO-1 and TuRBO-m highlights key algorithmic considerations central to this study. In the concurrent formulation, the acquisition/bandit strategies tend to prioritize changes to the MMC-related variables due to their stronger influence on the objective landscape, effectively biasing the search away from suitable LP configurations. This imbalance limits the optimizer’s ability to adequately explore and exploit the LP design space. Conversely, when the LP variables are optimized independently in the sequential setting, the algorithm can focus its modeling capacity and sampling budget exclusively on the LPs, leading to more effective exploration and improved configurations.

%In the concurrent setting, trust-region updates may be predominantly influenced by the topology parameters, causing trust-region-based methods to struggle with exploring certain areas of the lamination parameter subspace, particularly as the trust region becomes smaller. In this context, the sequential strategy proves more effective, as it allows for more thorough exploration of the lamination parameter subspace.

% For completeness, Figure~\ref{fig:box_plots_volume} presents the distribution of the volume constraint violations across all strategies and algorithms. As expected, when comparing Figures~\ref{fig:box_plots_volume} and~\ref{fig:box_plots_compliance}, the best-performing structures in terms of compliance are generally those with volume constraint values closest to zero. Notably, when comparing sequential and concurrent approaches, the sequential strategies tend to exhibit higher median values than their concurrent counterparts. This shows that since the volume constraint only applies to $\mathbf{x}_{\text{MMC}}$ and not the entire set of design variables, then the optimizer reaches larger feasible volumes.
 
% \begin{figure}[t]
%     \centering
% 	\includegraphics[width=0.9\textwidth]{./Figures/volume_constraint_full_budget.png}
%     \caption{Box plots of volume constraint distribution for each optimization algorithm under sequential and concurrent strategies.}
%     \label{fig:box_plots_volume}
% \end{figure}
% %
\subsection{Qualitative Comparison}
\label{subsec:qualitative_comparison}
In this section, we analyze the algorithms \mbox{TuRBO-1}, BAxUS, CMA-ES, and HEBO in terms of final material layout and LP distribution. The objective is to examine both the best-performing and median-performing structures, based on the evaluated modified compliance metric. The resulting fiber layouts, along with the distributions of $V_1$ and $V_3$, are presented in Tables~\ref{tab:concurrent_layouts} and~\ref{tab:sequential_layouts}.

\begin{table}[htbp]
\centering
\caption{Best-and median-performing structures for different algorithms using the concurrent setting.}
\label{tab:concurrent_layouts}
\resizebox{0.99\textwidth}{!}{%
\begin{tabular}{|l| c c | c c |}
\hline
\textbf{Algorithm} &
\multicolumn{2}{c|}{\textbf{Best-Performing Structure}} &
\multicolumn{2}{c|}{\textbf{Median-Performing Structure}} \\ \hline
 & \begin{small} $V_1$
\end{small}   & 
\begin{small} $V_3$ \end{small} & 
\begin{small}$V_1$ \end{small} & \begin{small} $V_3$ \end{small} \\
% CMA-ES row
\multirow{2}{*}{\textbf{CMA-ES}} &
\includegraphics[width=0.16\linewidth,trim={20cm 45cm 140cm 45cm},clip]{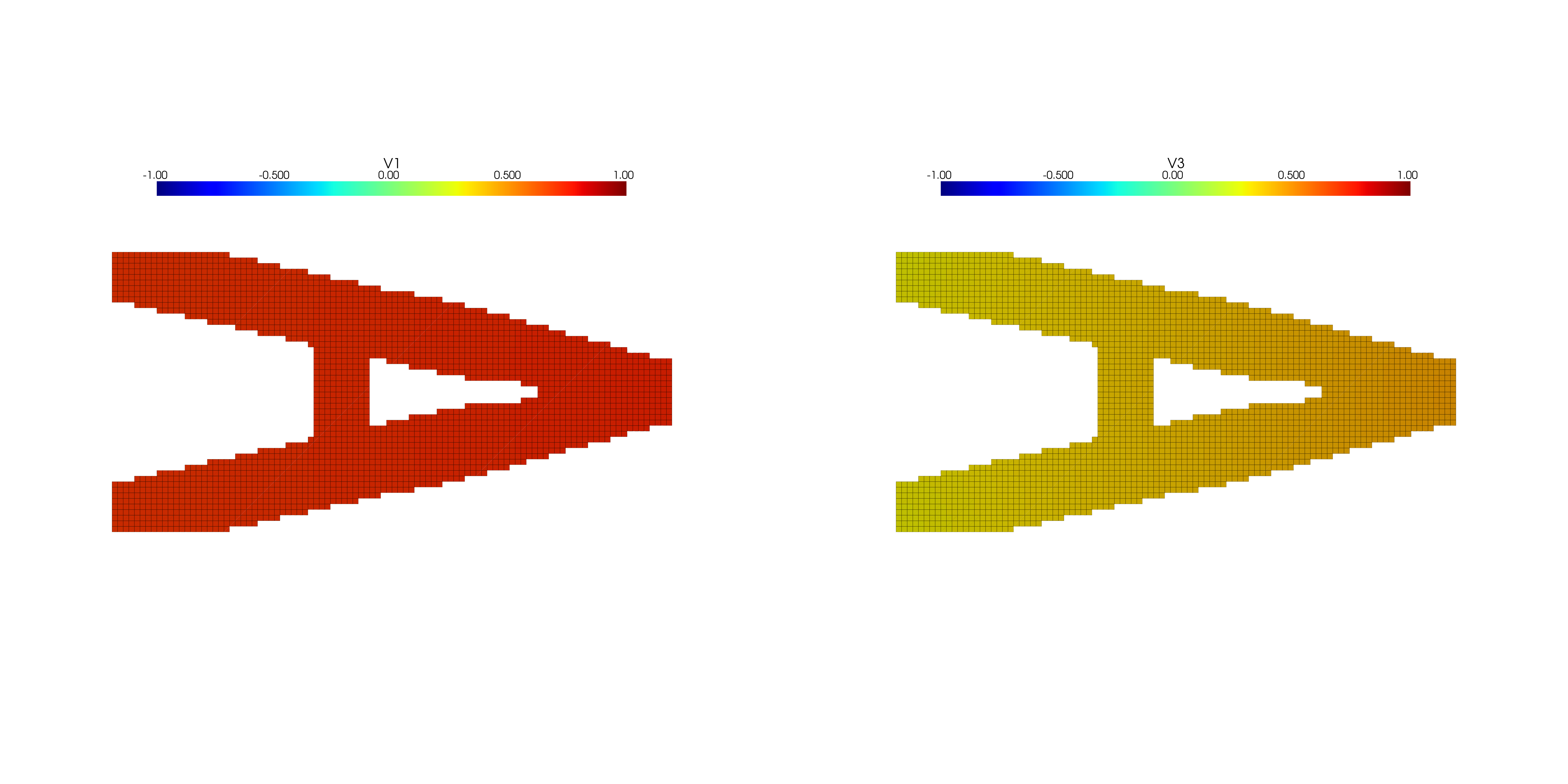} &
\includegraphics[width=0.16\linewidth,trim={140cm 45cm 20cm 45cm},clip]{./Figures/qualitative_comparison/Figures_Python/Run_8/best_lp_CMA_ES.png} &
\includegraphics[width=0.16\linewidth,trim={20cm 45cm 140cm 45cm},clip]{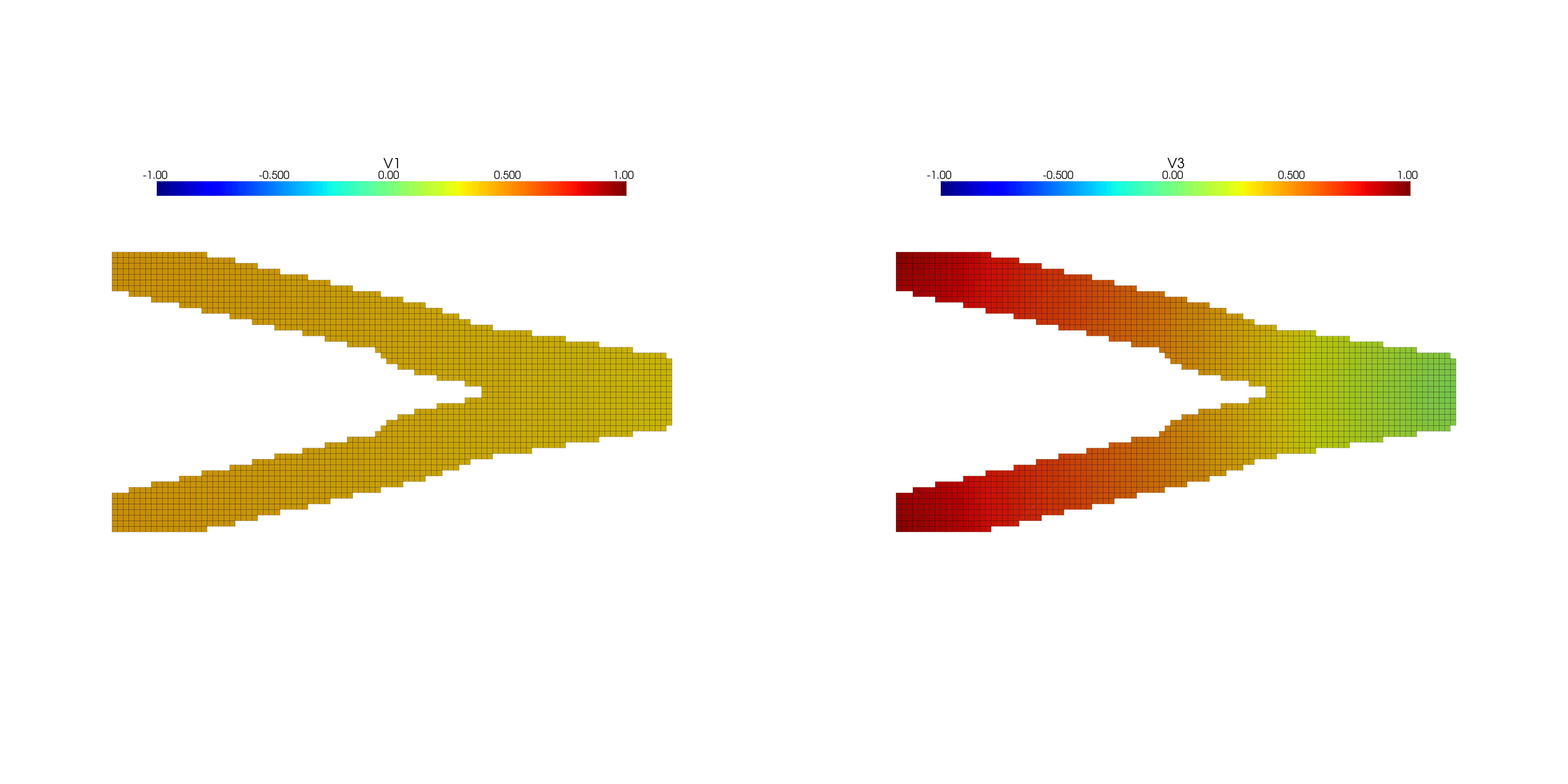} &
\includegraphics[width=0.16\linewidth,trim={140cm 45cm 20cm 45cm},clip]{./Figures/qualitative_comparison/Figures_Python/Run_5/median_lp_CMA_ES.png} \\
& \multicolumn{2}{c|}{\begin{small}$\tilde{C}=6.20583 \times 10^{-2}$ \end{small}} & \multicolumn{2}{c|}{\begin{small}$\tilde{C}=8.78738 \times 10^{-2}$ \end{small}} \\
% TuRBO-1 row
\multirow{2}{*}{\textbf{TuRBO-1}} &
\includegraphics[width=0.16\linewidth,trim={20cm 45cm 140cm 45cm},clip]{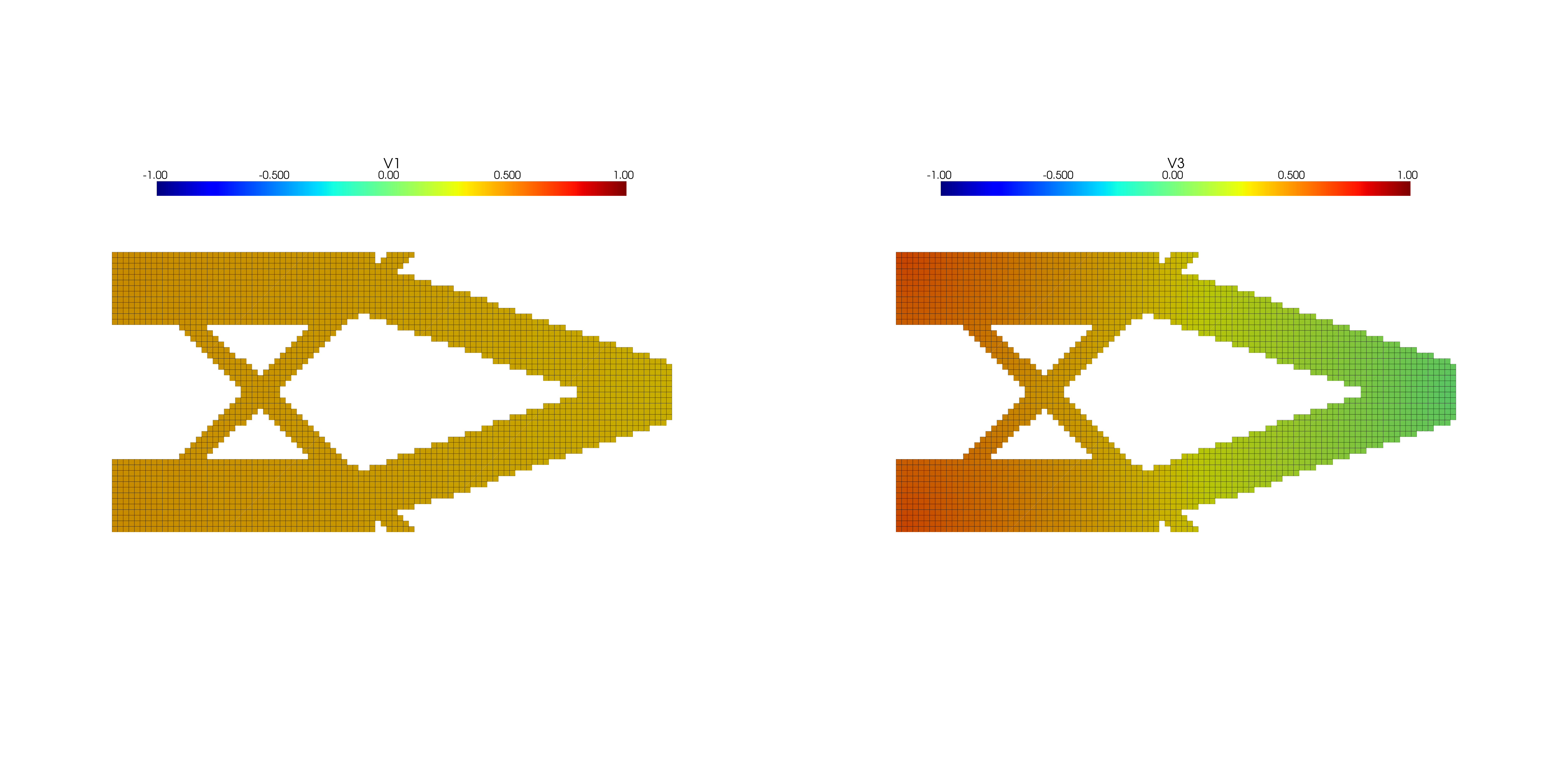} &
\includegraphics[width=0.16\linewidth,trim={140cm 45cm 20cm 45cm},clip]{./Figures/qualitative_comparison/Figures_Python/Run_26/best_lp_TuRBO-1.png} &
\includegraphics[width=0.16\linewidth,trim={20cm 45cm 140cm 45cm},clip]{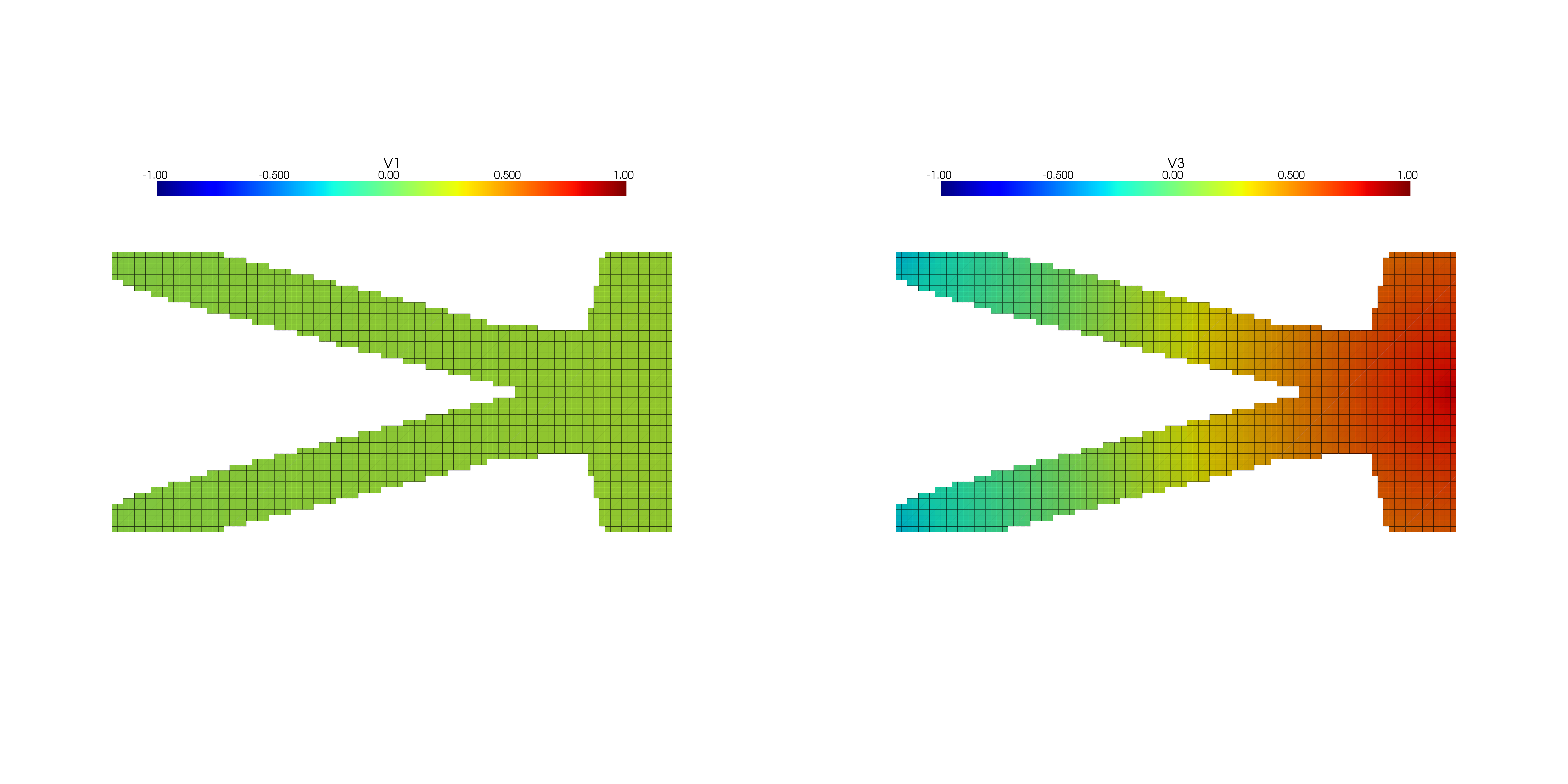} &
\includegraphics[width=0.16\linewidth,trim={140cm 45cm 20cm 45cm},clip]{./Figures/qualitative_comparison/Figures_Python/Run_37/median_lp_TuRBO-1.png} \\
& \multicolumn{2}{c|}{\begin{small}$\tilde{C}=6.00057 \times 10^{-2}$\end{small}} & \multicolumn{2}{c|}{\begin{small}$\tilde{C}=9.78883 \times 10^{-2}$\end{small}} \\
% HEBO row
\multirow{2}{*}{\textbf{HEBO}} &
\includegraphics[width=0.16\linewidth,trim={20cm 45cm 140cm 45cm},clip]{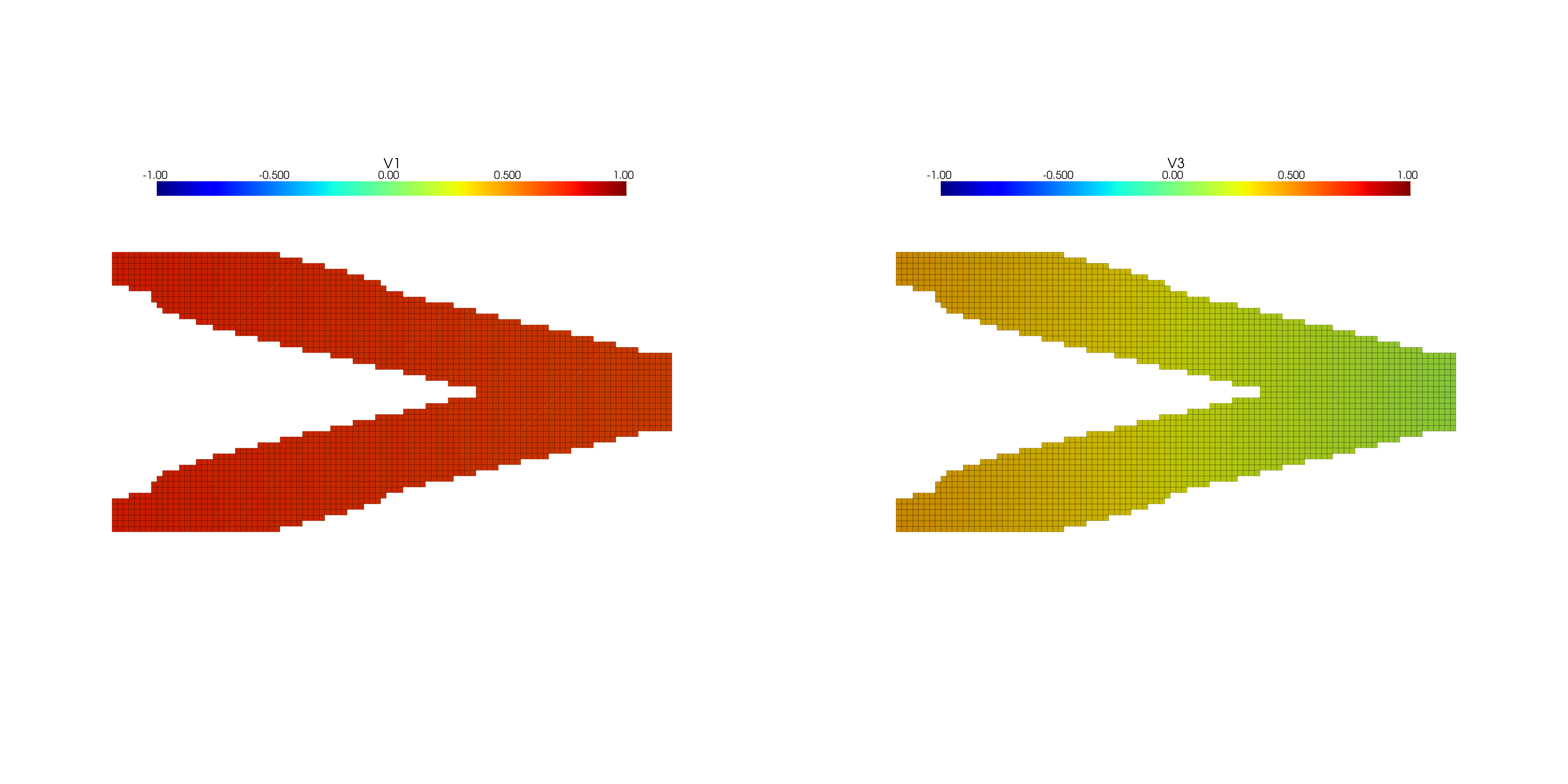} &
\includegraphics[width=0.16\linewidth,trim={140cm 45cm 20cm 45cm},clip]{./Figures/qualitative_comparison/Figures_Python/Run_127/best_lp_HEBO.png} &
\includegraphics[width=0.16\linewidth,trim={20cm 45cm 140cm 45cm},clip]{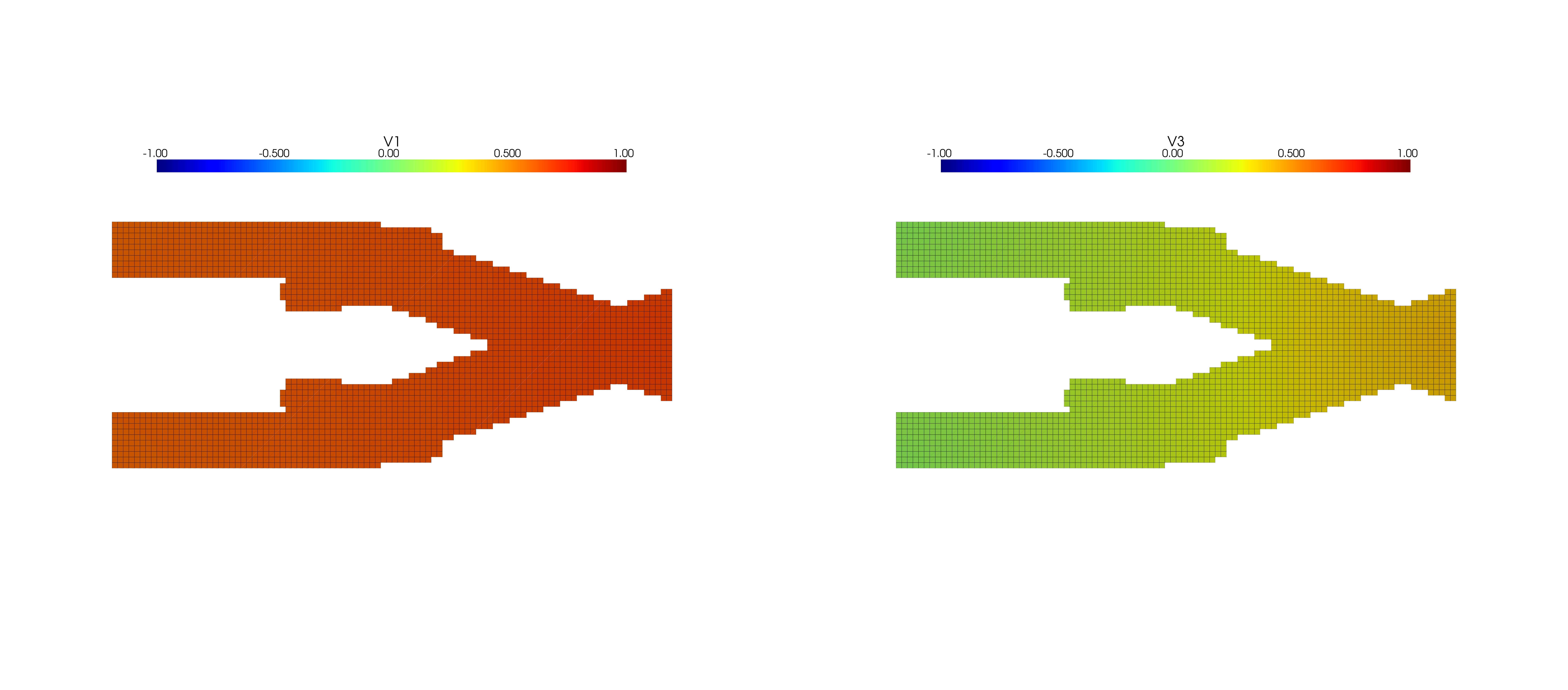} &
\includegraphics[width=0.16\linewidth,trim={140cm 45cm 20cm 45cm},clip]{./Figures/qualitative_comparison/Figures_Python/Run_130/median_lp_HEBO.png} \\
& \multicolumn{2}{c|}{\begin{small}$\tilde{C}=5.76630 \times 10^{-2}$ \end{small} } & \multicolumn{2}{c|}{\begin{small}$\tilde{C}=8.82539 \times 10^{-2}$\end{small} } \\
% BAxUS row
\multirow{2}{*}{\textbf{BAxUS}} &
\includegraphics[width=0.16\linewidth,trim={20cm 45cm 140cm 45cm},clip]{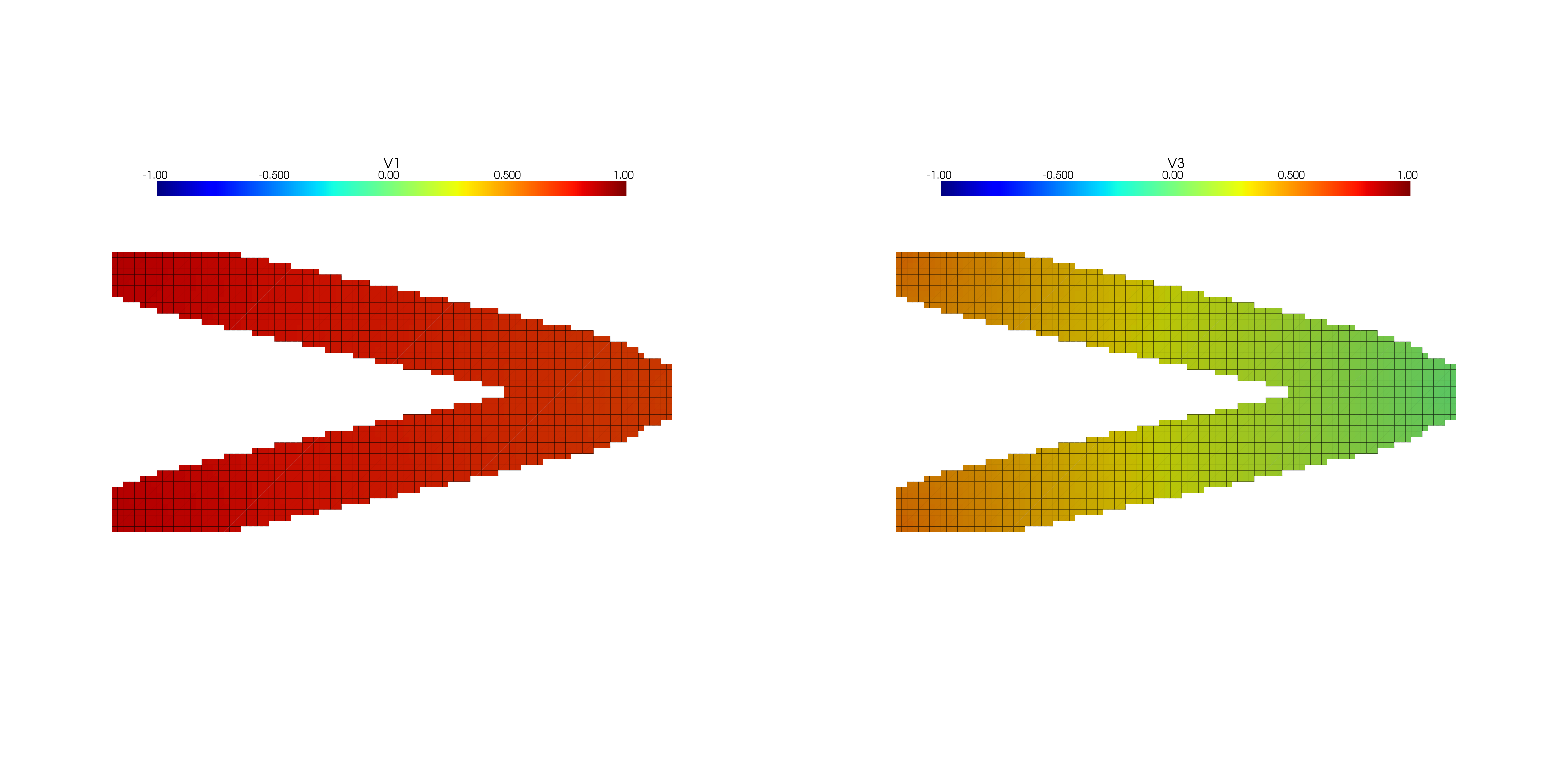} &
\includegraphics[width=0.16\linewidth,trim={140cm 45cm 20cm 45cm},clip]{./Figures/qualitative_comparison/Figures_Python/Run_117/best_lp_BAxUS.png} &
\includegraphics[width=0.16\linewidth,trim={20cm 45cm 140cm 45cm},clip]{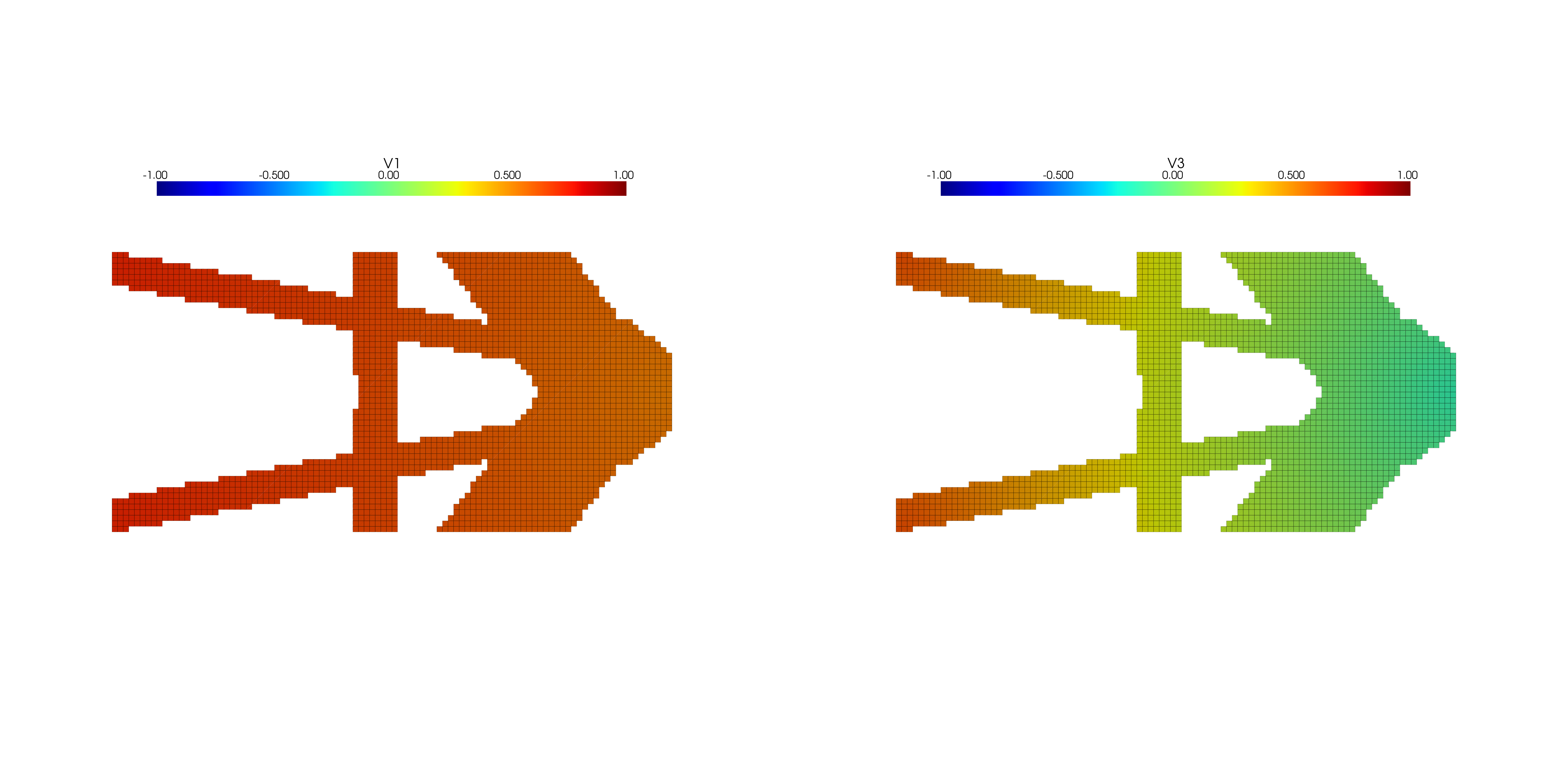} &
\includegraphics[width=0.16\linewidth,trim={140cm 45cm 20cm 45cm},clip]{./Figures/qualitative_comparison/Figures_Python/Run_104/median_lp_BAxUS.png} \\
& \multicolumn{2}{c|}{\begin{small}$\tilde{C}=5.75949 \times 10^{-2}$\end{small}} & \multicolumn{2}{c|}{\begin{small}$\tilde{C}=1.11891 \times 10^{-1}$ \end{small} } \\
& \multicolumn{2}{c}{} & \multicolumn{2}{c|}{} \\
& \multicolumn{4}{c|}{\def\svgwidth{0.45\linewidth}
%% Creator: Inkscape 1.3.2 (091e20e, 2023-11-25, custom), www.inkscape.org
%% PDF/EPS/PS + LaTeX output extension by Johan Engelen, 2010
%% Accompanies image file 'colorbar_v1_v3_distributions.pdf' (pdf, eps, ps)
%%
%% To include the image in your LaTeX document, write
%%   \input{<filename>.pdf_tex}
%%  instead of
%%   \includegraphics{<filename>.pdf}
%% To scale the image, write
%%   \def\svgwidth{<desired width>}
%%   \input{<filename>.pdf_tex}
%%  instead of
%%   \includegraphics[width=<desired width>]{<filename>.pdf}
%%
%% Images with a different path to the parent latex file can
%% be accessed with the `import' package (which may need to be
%% installed) using
%%   \usepackage{import}
%% in the preamble, and then including the image with
%%   \import{<path to file>}{<filename>.pdf_tex}
%% Alternatively, one can specify
%%   \graphicspath{{<path to file>/}}
%% 
%% For more information, please see info/svg-inkscape on CTAN:
%%   http://tug.ctan.org/tex-archive/info/svg-inkscape
%%
\begingroup%
  \makeatletter%
  \providecommand\color[2][]{%
    \errmessage{(Inkscape) Color is used for the text in Inkscape, but the package 'color.sty' is not loaded}%
    \renewcommand\color[2][]{}%
  }%
  \providecommand\transparent[1]{%
    \errmessage{(Inkscape) Transparency is used (non-zero) for the text in Inkscape, but the package 'transparent.sty' is not loaded}%
    \renewcommand\transparent[1]{}%
  }%
  \providecommand\rotatebox[2]{#2}%
  \newcommand*\fsize{\dimexpr\f@size pt\relax}%
  \newcommand*\lineheight[1]{\fontsize{\fsize}{#1\fsize}\selectfont}%
  \ifx\svgwidth\undefined%
    \setlength{\unitlength}{1960.33813477bp}%
    \ifx\svgscale\undefined%
      \relax%
    \else%
      \setlength{\unitlength}{\unitlength * \real{\svgscale}}%
    \fi%
  \else%
    \setlength{\unitlength}{\svgwidth}%
  \fi%
  \global\let\svgwidth\undefined%
  \global\let\svgscale\undefined%
  \makeatother%
  \begin{picture}(1,0.0517682)%
    \lineheight{1}%
    \setlength\tabcolsep{0pt}%
    \put(0.00,0.03401618){\color[rgb]{0,0,0}\makebox(0,0)[ct]{\lineheight{1.25}\smash{\begin{tabular}[t]{l}-1.000\end{tabular}}}}%
    \put(1.00,0.03401618){\color[rgb]{0,0,0}\makebox(0,0)[ct]{\lineheight{1.25}\smash{\begin{tabular}[t]{l}1.000\end{tabular}}}}%
    \put(0.50,0.03401618){\color[rgb]{0,0,0}\makebox(0,0)[ct]{\lineheight{1.25}\smash{\begin{tabular}[t]{l}0.000\end{tabular}}}}%
    \put(0.25,0.03401618){\color[rgb]{0,0,0}\makebox(0,0)[ct]{\lineheight{1.25}\smash{\begin{tabular}[t]{l}-0.500\end{tabular}}}}%
    \put(0.75,0.03401618){\color[rgb]{0,0,0}\makebox(0,0)[ct]{\lineheight{1.25}\smash{\begin{tabular}[t]{l}0.500\end{tabular}}}}%
    \put(0,0){\includegraphics[width=\unitlength,page=1]{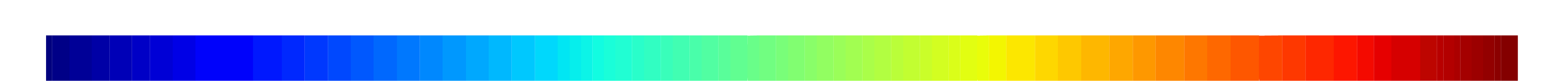}}%
  \end{picture}%
\endgroup%
}  \\
\hline
\end{tabular}%
}
\end{table}
\begin{table}[htbp]
\centering
\caption{Best- and median-performing structures for different algorithms using the sequential setting.}
\label{tab:sequential_layouts}
\resizebox{0.99\textwidth}{!}{%
\begin{tabular}{|l|cc|cc|}
\hline
\textbf{Algorithm} & \multicolumn{2}{c|}{\textbf{Best-Performing Structure}}  & \multicolumn{2}{c|}{\textbf{Median-Performing Structure}} \\ \hline
 & \begin{small} $V_1$
\end{small}   & 
\begin{small} $V_3$ \end{small} & 
\begin{small}$V_1$ \end{small} & \begin{small} $V_3$ \end{small} \\
\multirow{2}{*}{\textbf{CMA-ES}} &
  \includegraphics[width=0.16\linewidth,trim={20cm 45cm 140cm 45cm},clip]{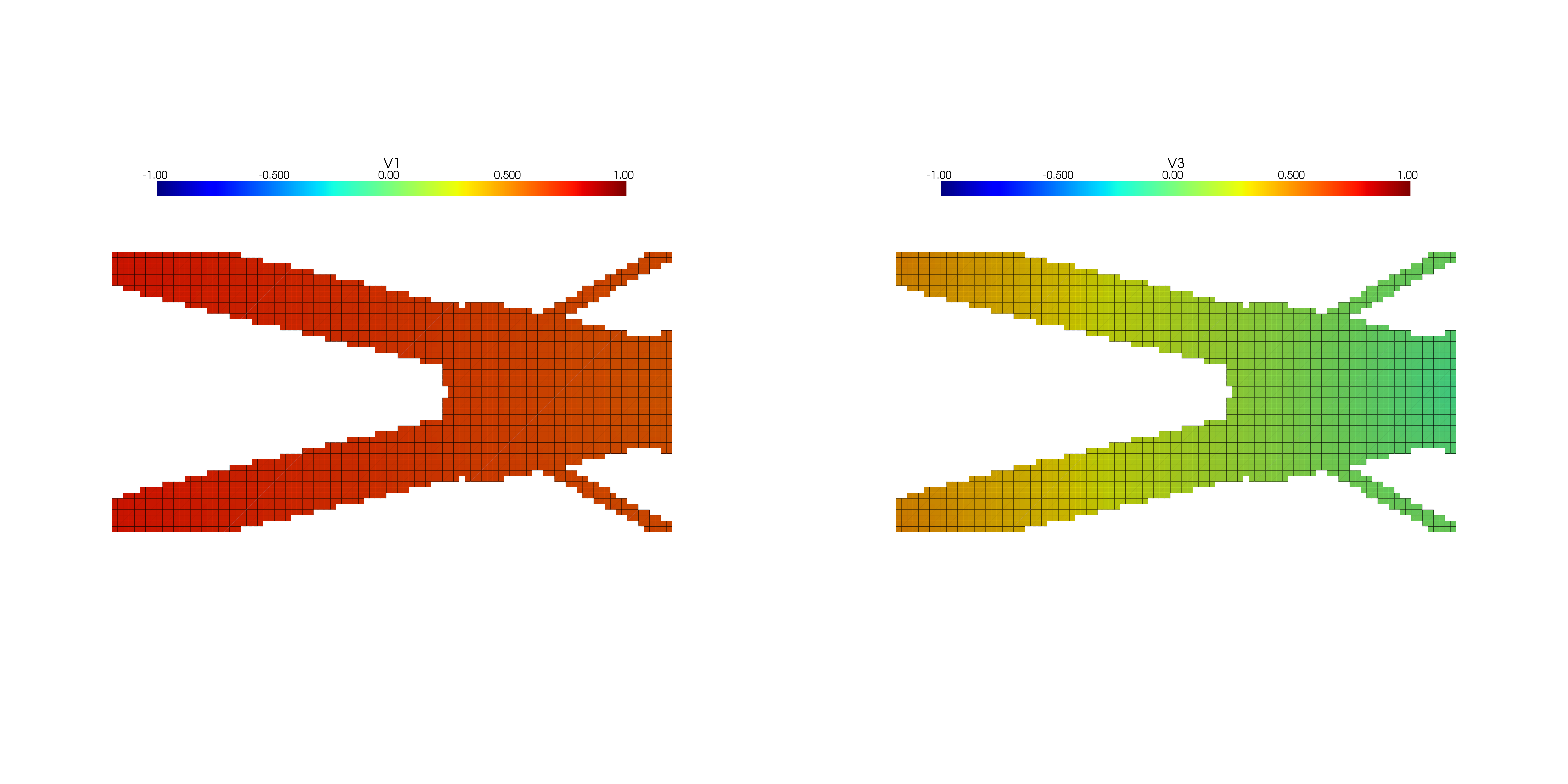} &
  \includegraphics[width=0.16\linewidth,trim={140cm 45cm 20cm 45cm},clip]{./Figures/qualitative_comparison/Figures_Python/Run_2145/best_lp_CMA-ES.png} &
  \includegraphics[width=0.16\linewidth,trim={20cm 45cm 140cm 45cm},clip]{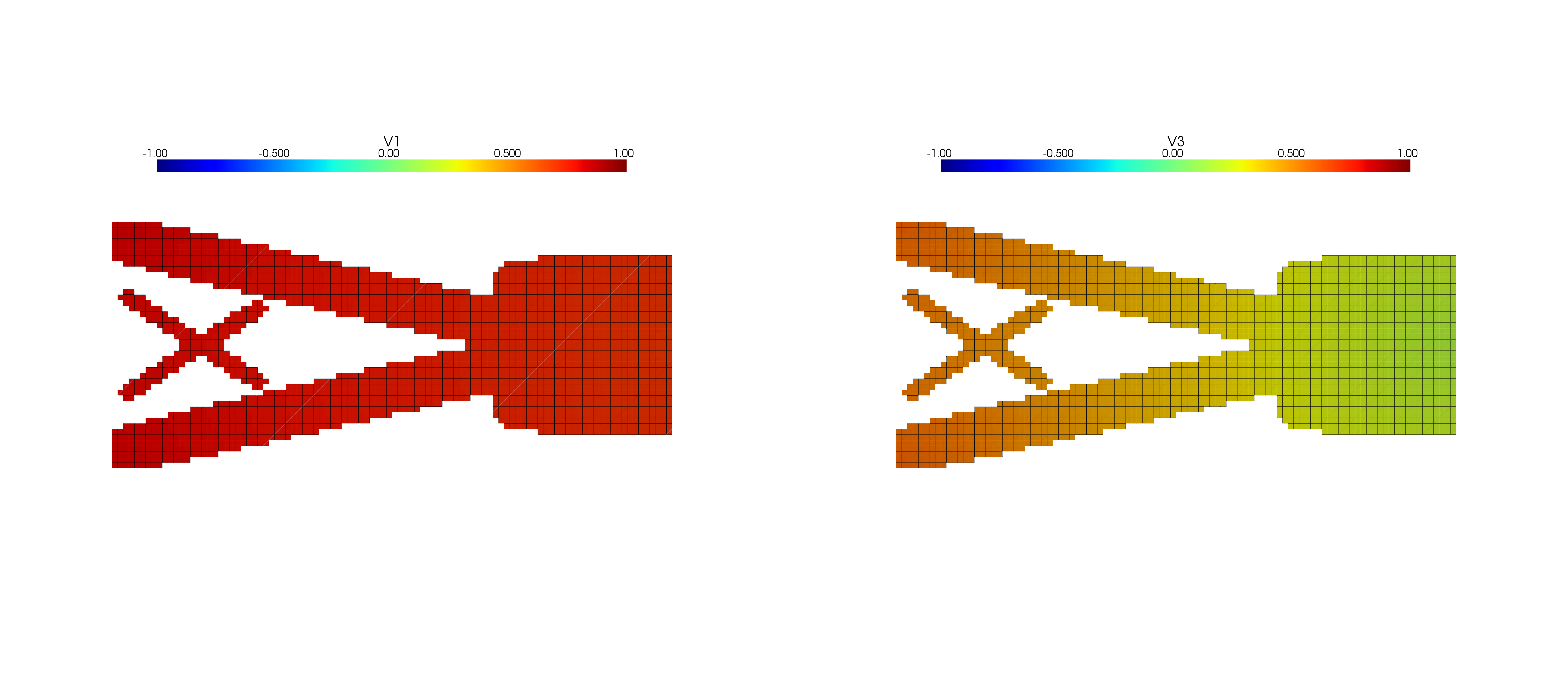} &
  \includegraphics[width=0.16\linewidth,trim={140cm 45cm 20cm 45cm},clip]{./Figures/qualitative_comparison/Figures_Python/Run_2169/median_lp_CMA-ES.png} \\
                   & \multicolumn{2}{c|}{ \begin{small}$\tilde{C}=6.17188 \times 10^{-2} $ \end{small}} & \multicolumn{2}{c|}{\begin{small}$\tilde{C}=1.09087 \times 10^{-1}$ \end{small} }   \\
\multirow{2}{*}{\textbf{TuRBO-1}} &
  \includegraphics[width=0.16\linewidth,trim={20cm 45cm 140cm 45cm},clip]{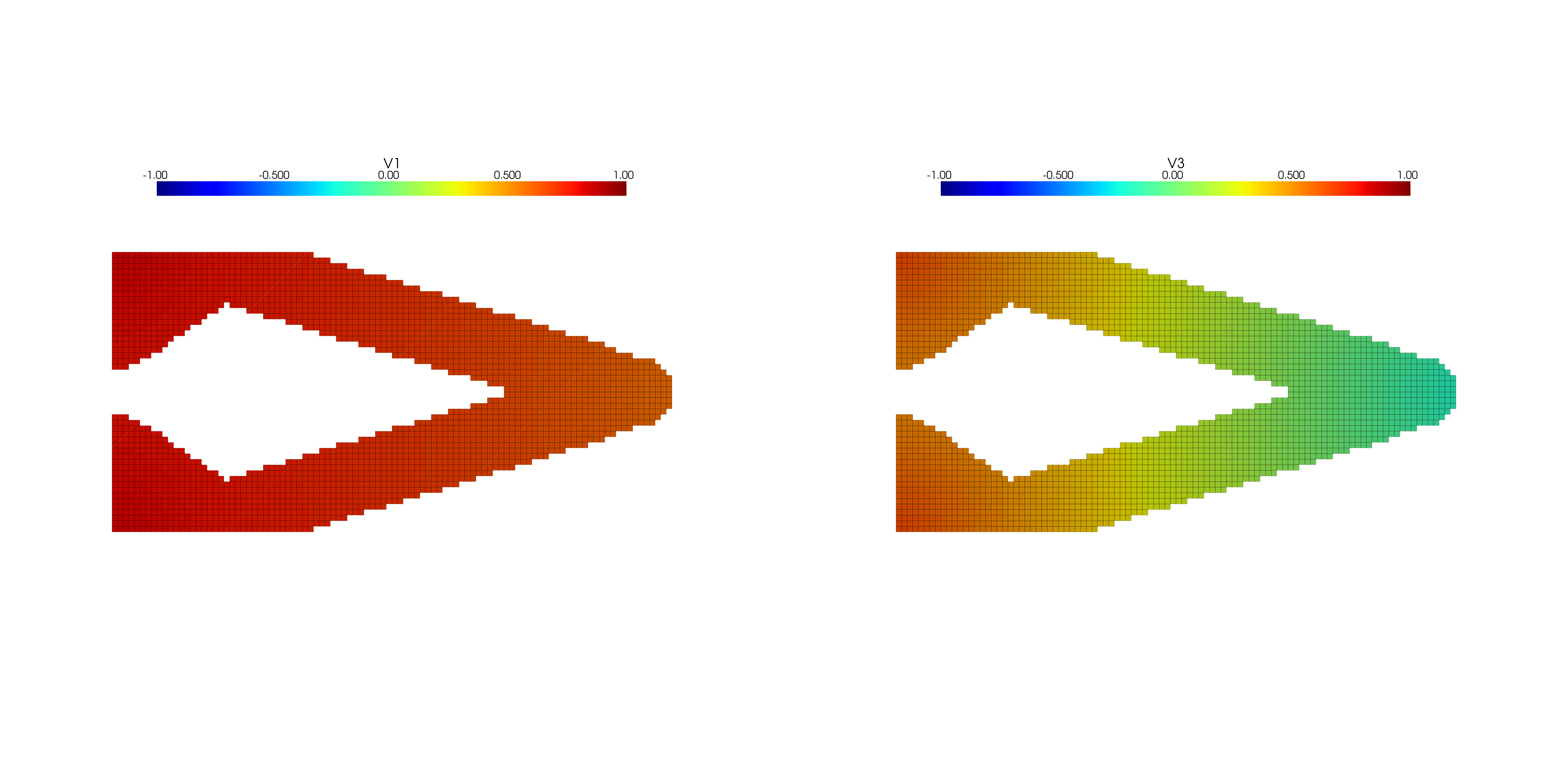} &
  \includegraphics[width=0.16\linewidth,trim={140cm 45cm 20cm 45cm},clip]{./Figures/qualitative_comparison/Figures_Python/Run_2211/best_lp_TuRBO-1.png} &
  \includegraphics[width=0.16\linewidth,trim={20cm 45cm 140cm 45cm},clip]{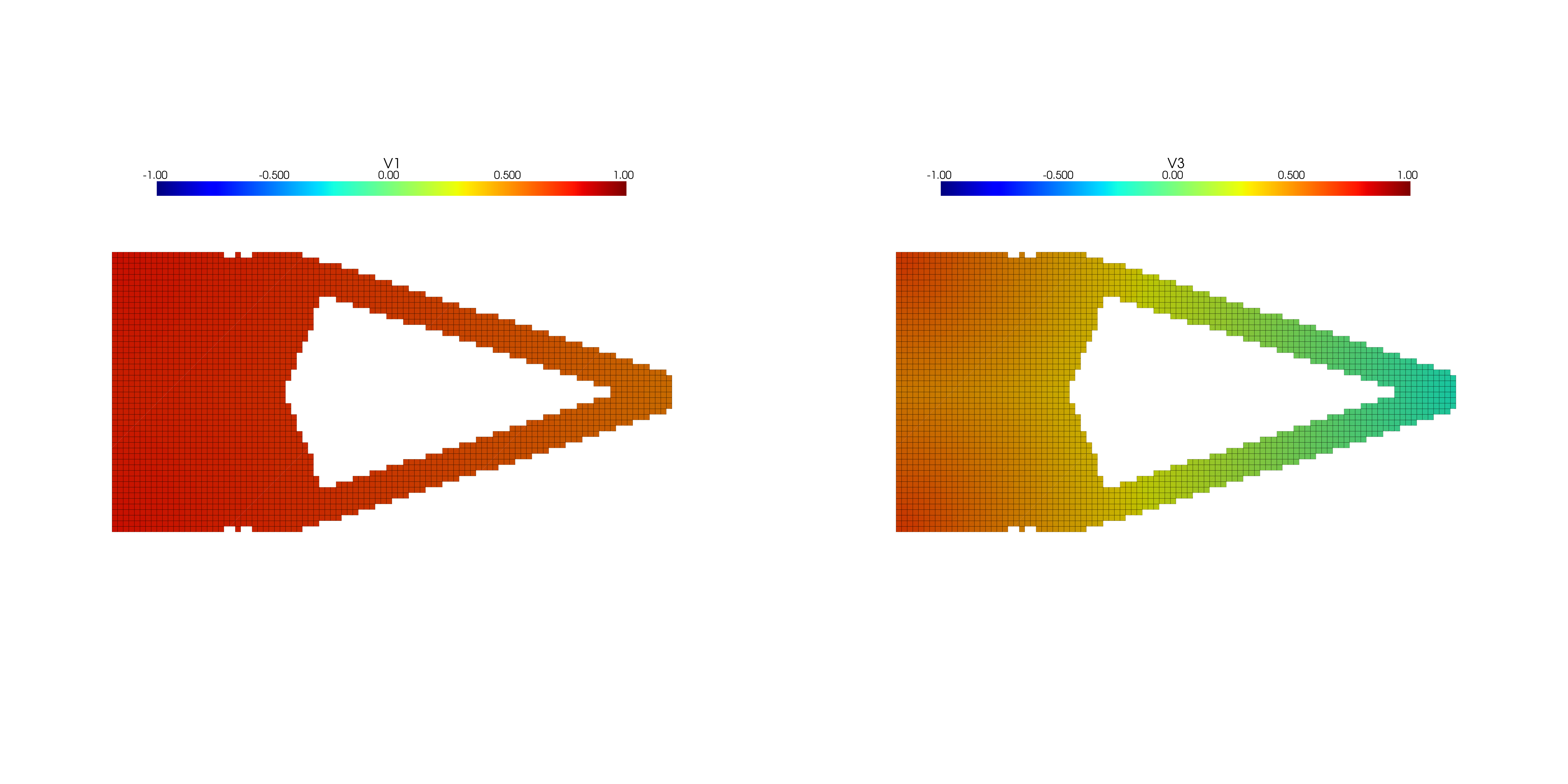} &
  \includegraphics[width=0.16\linewidth,trim={140cm 45cm 20cm 45cm},clip]{./Figures/qualitative_comparison/Figures_Python/Run_2201/median_lp_TuRBO-1.png} \\
                   & \multicolumn{2}{c|}{\begin{small}$\tilde{C}=5.79513 \times 10^{-2} $ \end{small}} & \multicolumn{2}{c|}{\begin{small}$\tilde{C}=7.13461 \times 10^{-2}$ \end{small}}   \\
\multirow{2}{*}{\textbf{HEBO}} &
  \includegraphics[width=0.16\linewidth,trim={20cm 45cm 140cm 45cm},clip]{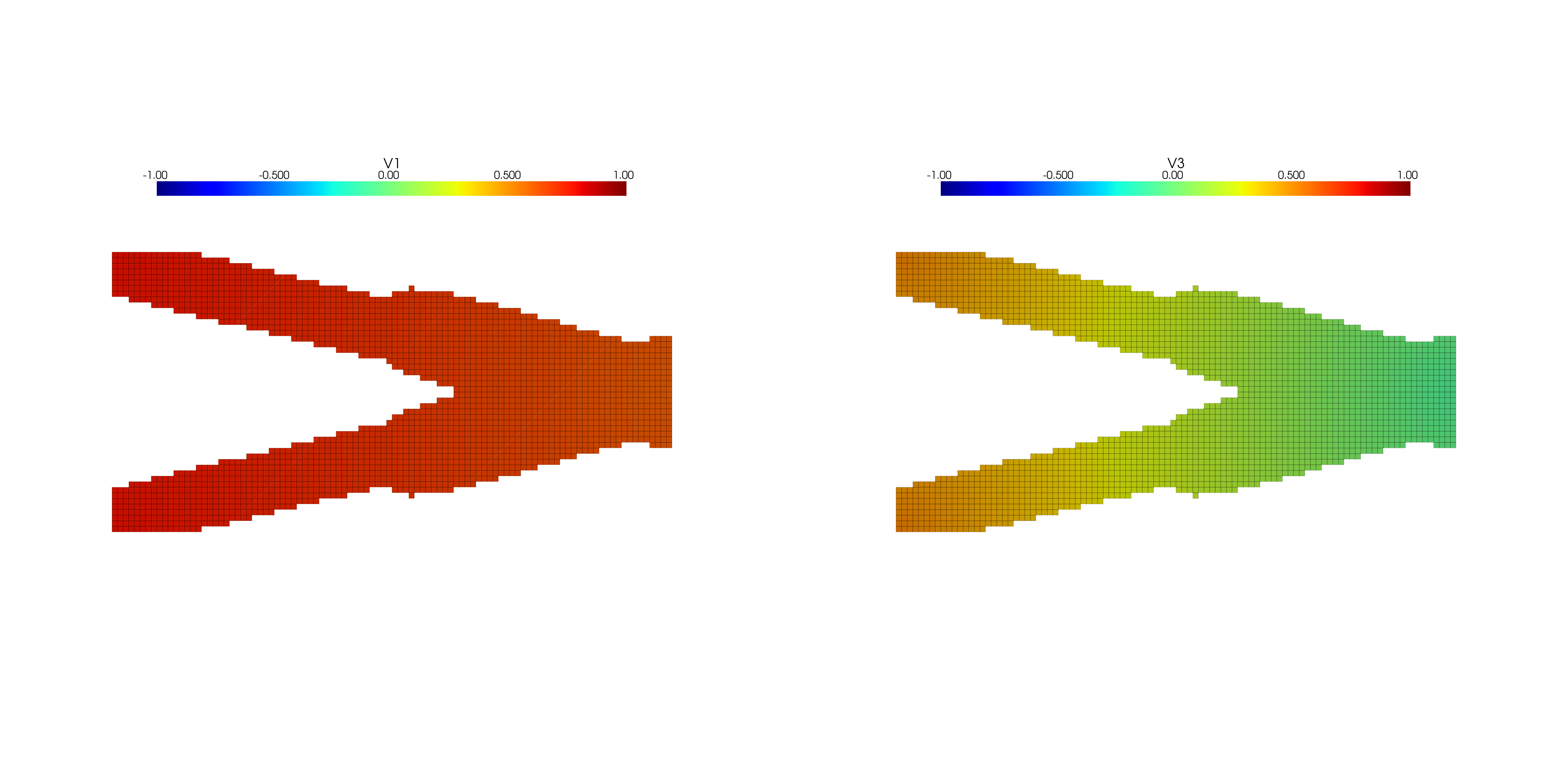} &
  \includegraphics[width=0.16\linewidth,trim={140cm 45cm 20cm 45cm},clip]{./Figures/qualitative_comparison/Figures_Python/Run_2403/best_lp_HEBO.png} &
  \includegraphics[width=0.16\linewidth,trim={20cm 45cm 140cm 45cm},clip]{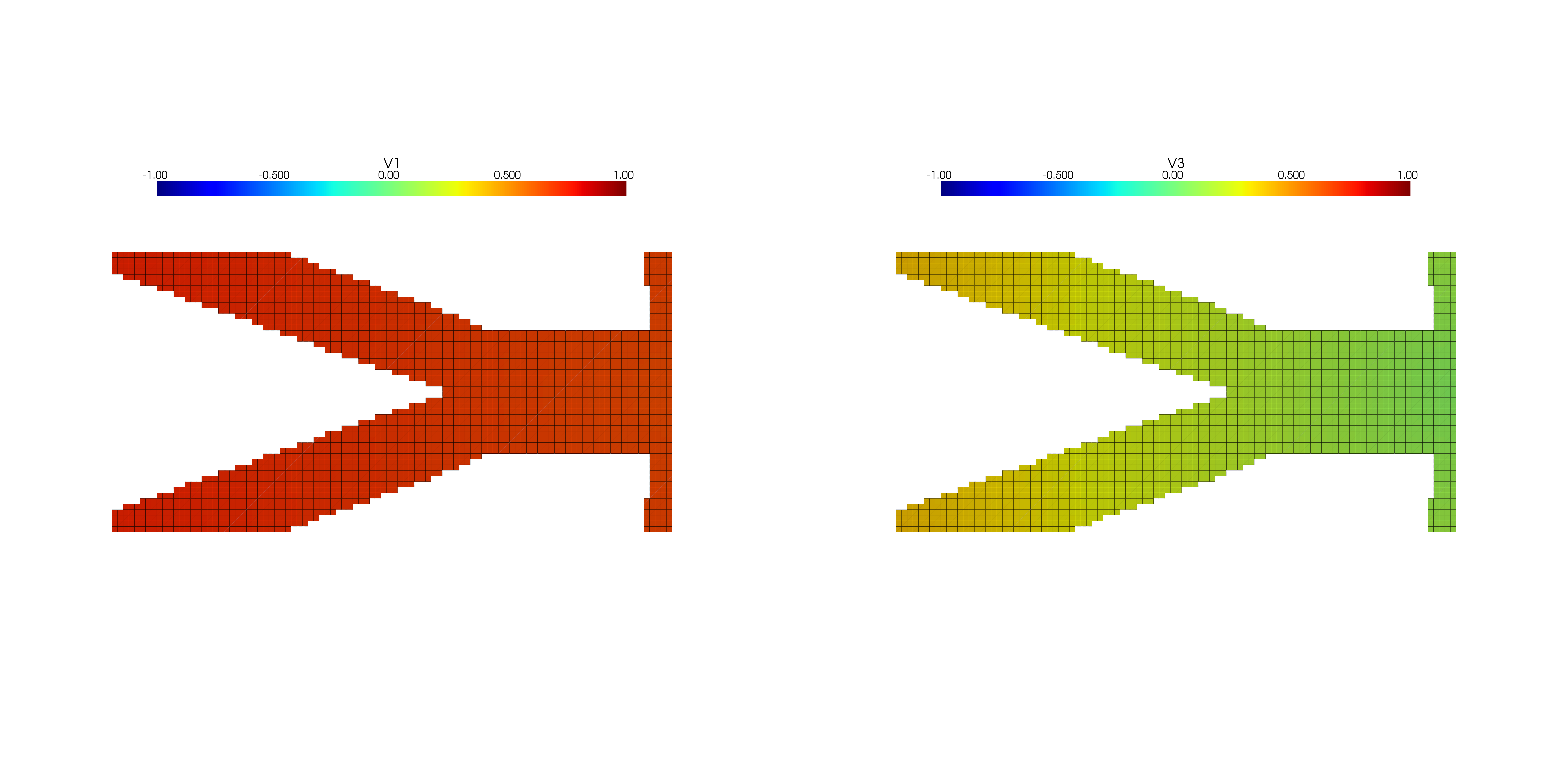} &
  \includegraphics[width=0.16\linewidth,trim={140cm 45cm 20cm 45cm},clip]{./Figures/qualitative_comparison/Figures_Python/Run_2417/median_lp_HEBO.png} \\
                   & \multicolumn{2}{c|}{\begin{small}$\tilde{C}=5.83527 \times 10^{-2}$ \end{small} } & \multicolumn{2}{c|}{\begin{small}$\tilde{C}=6.91426 \times 10^{-2}$ \end{small}}   \\
\multirow{2}{*}{\textbf{BAxUS}} &
  \includegraphics[width=0.16\linewidth,trim={20cm 45cm 140cm 45cm},clip]{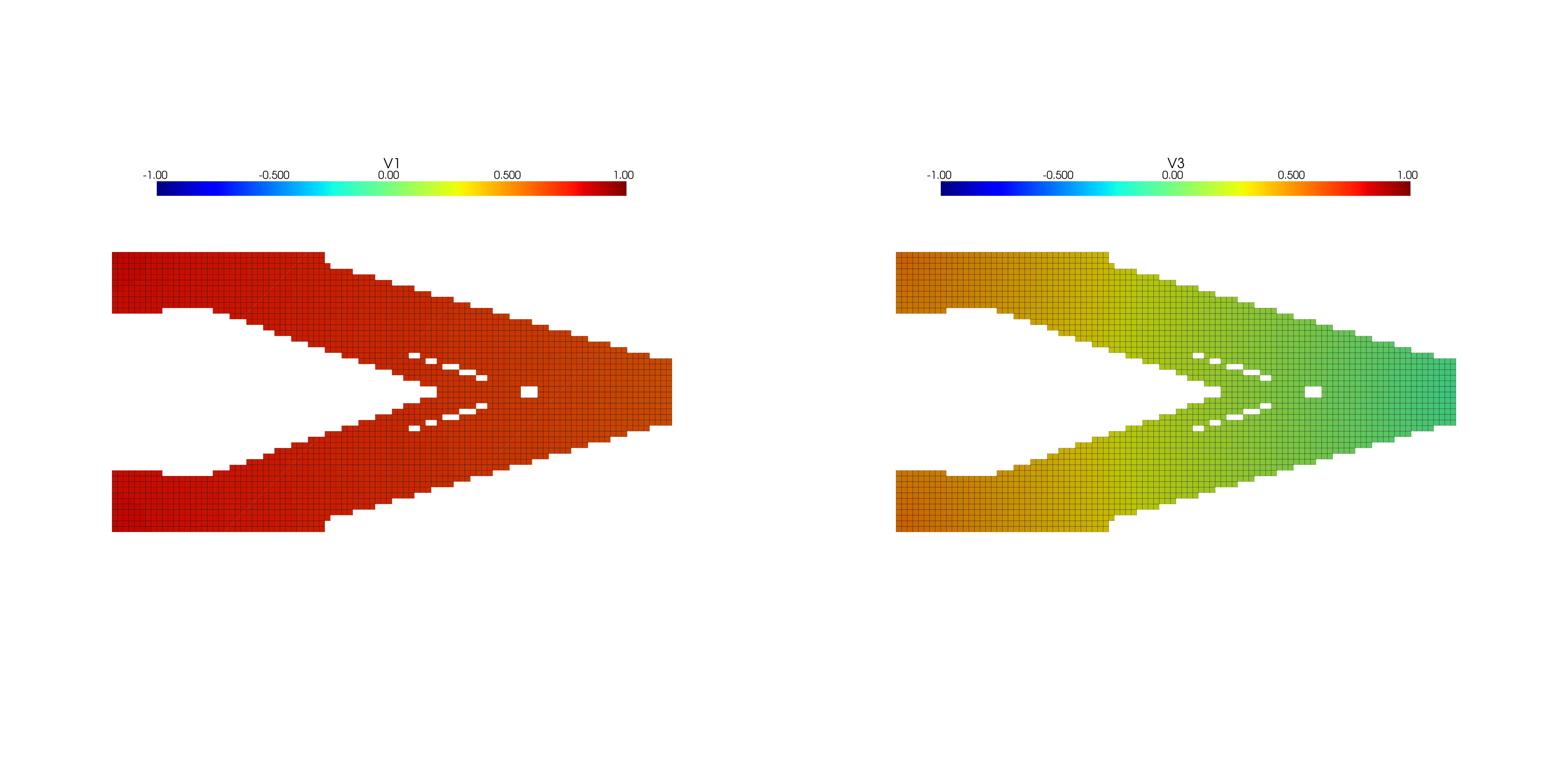} &
  \includegraphics[width=0.16\linewidth,trim={140cm 45cm 20cm 45cm},clip]{./Figures/qualitative_comparison/Figures_Python/Run_2373/best_lp_BAxUS.png} &
  \includegraphics[width=0.16\linewidth,trim={20cm 45cm 140cm 45cm},clip]{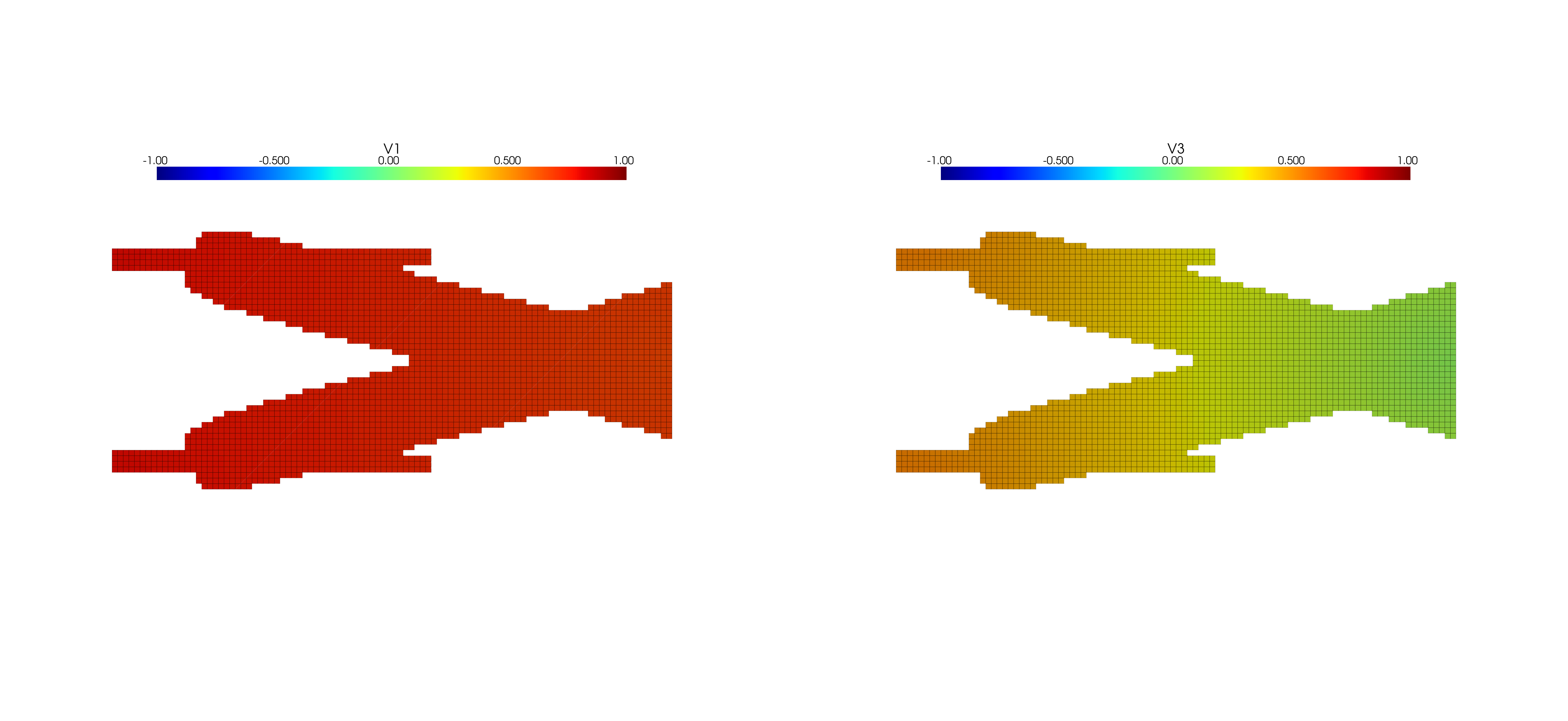} &
  \includegraphics[width=0.16\linewidth,trim={140cm 45cm 20cm 45cm},clip]{./Figures/qualitative_comparison/Figures_Python/Run_2343/median_lp_BAxUS.png} \\
                   & \multicolumn{2}{c|}{\begin{small}$\tilde{C}=5.70230 \times 10^{-2}$ \end{small}} & \multicolumn{2}{c|}{\begin{small}$\tilde{C}=1.06861 \times 10^{-1}$\end{small}}   \\
                   & \multicolumn{2}{c}{} & \multicolumn{2}{c|}{}   \\
                   & \multicolumn{4}{c|}{\def\svgwidth{0.45\linewidth} }        \\ \hline
\end{tabular}%
}
\end{table}

\begin{figure}[htbp]
    \centering
    \def\svgwidth{0.95\linewidth}
    %% Creator: Inkscape 1.3.2 (091e20e, 2023-11-25, custom), www.inkscape.org
%% PDF/EPS/PS + LaTeX output extension by Johan Engelen, 2010
%% Accompanies image file 'fiber_angle_distribution_3.pdf' (pdf, eps, ps)
%%
%% To include the image in your LaTeX document, write
%%   \input{<filename>.pdf_tex}
%%  instead of
%%   \includegraphics{<filename>.pdf}
%% To scale the image, write
%%   \def\svgwidth{<desired width>}
%%   \input{<filename>.pdf_tex}
%%  instead of
%%   \includegraphics[width=<desired width>]{<filename>.pdf}
%%
%% Images with a different path to the parent latex file can
%% be accessed with the `import' package (which may need to be
%% installed) using
%%   \usepackage{import}
%% in the preamble, and then including the image with
%%   \import{<path to file>}{<filename>.pdf_tex}
%% Alternatively, one can specify
%%   \graphicspath{{<path to file>/}}
%% 
%% For more information, please see info/svg-inkscape on CTAN:
%%   http://tug.ctan.org/tex-archive/info/svg-inkscape
%%
\begingroup%
  \makeatletter%
  \providecommand\color[2][]{%
    \errmessage{(Inkscape) Color is used for the text in Inkscape, but the package 'color.sty' is not loaded}%
    \renewcommand\color[2][]{}%
  }%
  \providecommand\transparent[1]{%
    \errmessage{(Inkscape) Transparency is used (non-zero) for the text in Inkscape, but the package 'transparent.sty' is not loaded}%
    \renewcommand\transparent[1]{}%
  }%
  \providecommand\rotatebox[2]{#2}%
  \newcommand*\fsize{\dimexpr\f@size pt\relax}%
  \newcommand*\lineheight[1]{\fontsize{\fsize}{#1\fsize}\selectfont}%
  \ifx\svgwidth\undefined%
    \setlength{\unitlength}{5851.21472168bp}%
    \ifx\svgscale\undefined%
      \relax%
    \else%
      \setlength{\unitlength}{\unitlength * \real{\svgscale}}%
    \fi%
  \else%
    \setlength{\unitlength}{\svgwidth}%
  \fi%
  \global\let\svgwidth\undefined%
  \global\let\svgscale\undefined%
  \makeatother%
  \begin{picture}(1,0.26519524)%
    \lineheight{1}%
    \setlength\tabcolsep{0pt}%
    \put(0,0){\includegraphics[width=\unitlength,page=1]{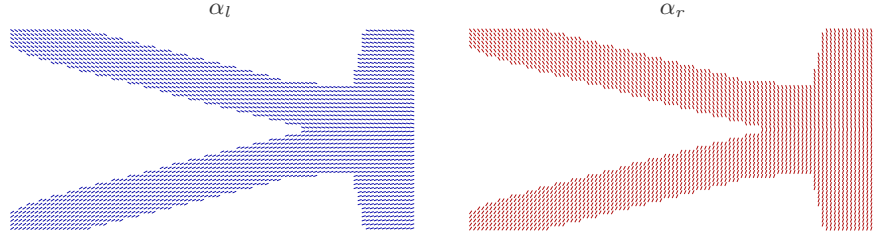}}%
    \put(0.23680722,0.25380946){\color[rgb]{0.2,0.2,0.2}\makebox(0,0)[lt]{\lineheight{1.25}\smash{\begin{tabular}[t]{l}$\alpha_{l}$\end{tabular}}}}%
    \put(0.75376128,0.25380946){\color[rgb]{0.2,0.2,0.2}\makebox(0,0)[lt]{\lineheight{1.25}\smash{\begin{tabular}[t]{l}$\alpha_{r}$\end{tabular}}}}%
  \end{picture}%
\endgroup%

    \caption{Fiber angle distributions for the median performing structure obtained with TuRBO-1 algorithm with the concurrent strategy. The fiber orientation distributions $\alpha_l$ and $\alpha_r$ correspond to the distributions by setting $R_l=1$ and $R_r=1$, respectively. In this case $R_r=0.582$.}
    \label{fig:svg_plot_1}
\end{figure}

From these tables, it can be observed that the distributions of \(V_1\) and \(V_3\) for each algorithm under the sequential approach are highly consistent. A similar LP distribution is also observed among the best performing structures obtained with concurrent strategies, particularly for TuRBO-1, HEBO, and BAxUS. However, when examining the median-performing structures under the concurrent approach, the LP distributions appear more diverse, suggesting a broader range of LP configurations in these cases. 
%
% This diversity may be attributed to convergence to local minima driven by suboptimal topologies, where the optimizer adjusts the LPs in response. 
The observed diversity in performance may stem from the dominant role of topology parameters in driving convergence, as mentioned above. Since material layout primarily influences modified compliance, adjustments in LPs often act as reactive fine-tuning steps rather than substantial drivers of performance, particularly when the underlying topology is suboptimal.
An illustrative example is the median performance structure obtained using TuRBO-1 in a concurrent setting. In this case, the \(V_3\) distribution appears inverted compared to that observed in the sequential cases. To provide further insight, Figure~\ref{fig:svg_plot_1} presents the corresponding fiber angle distribution. 
%
%As shown in the figure, the fibers near the walls (left side) exhibit similar orientations on both sides, whereas near the load application region (right side), the fibers are oriented almost orthogonally. While these fiber orientations may be considered suboptimal from a global perspective, the local configuration—specifically the presence of nearly vertical material near the load—makes such orientation reasonable. Nevertheless, this result also suggests that excess material is present in unstressed regions, contributing little to none load-bearing and potentially reducing structural efficiency.
%
As shown in the figure, both \(\alpha_l\) and \(\alpha_r\) display similar fiber orientations near the clamped boundary. In contrast, near the load application point, \(\alpha_l\) aligns predominantly with the horizontal axis, while \(\alpha_r\) aligns with the vertical axis. This difference suggests that for the given topology, the local principal stress directions are aligned with the coordinate axes. This behavior is consistent with the presence of a vertical strip of material in a region where the bending moments are smaller than the axial loads. However, there is also a noticeable presence of excess material in regions subjected to negligible stress. Such formations contribute minimally to the load-bearing capacity and have a negative impact on the overall structural efficiency.
\begin{figure}[htbp]
    \centering
    \def\svgwidth{0.95\linewidth}
    %% Creator: Inkscape 1.3.2 (091e20e, 2023-11-25, custom), www.inkscape.org
%% PDF/EPS/PS + LaTeX output extension by Johan Engelen, 2010
%% Accompanies image file 'fiber_angle_distribution_4.pdf' (pdf, eps, ps)
%%
%% To include the image in your LaTeX document, write
%%   \input{<filename>.pdf_tex}
%%  instead of
%%   \includegraphics{<filename>.pdf}
%% To scale the image, write
%%   \def\svgwidth{<desired width>}
%%   \input{<filename>.pdf_tex}
%%  instead of
%%   \includegraphics[width=<desired width>]{<filename>.pdf}
%%
%% Images with a different path to the parent latex file can
%% be accessed with the `import' package (which may need to be
%% installed) using
%%   \usepackage{import}
%% in the preamble, and then including the image with
%%   \import{<path to file>}{<filename>.pdf_tex}
%% Alternatively, one can specify
%%   \graphicspath{{<path to file>/}}
%% 
%% For more information, please see info/svg-inkscape on CTAN:
%%   http://tug.ctan.org/tex-archive/info/svg-inkscape
%%
\begingroup%
  \makeatletter%
  \providecommand\color[2][]{%
    \errmessage{(Inkscape) Color is used for the text in Inkscape, but the package 'color.sty' is not loaded}%
    \renewcommand\color[2][]{}%
  }%
  \providecommand\transparent[1]{%
    \errmessage{(Inkscape) Transparency is used (non-zero) for the text in Inkscape, but the package 'transparent.sty' is not loaded}%
    \renewcommand\transparent[1]{}%
  }%
  \providecommand\rotatebox[2]{#2}%
  \newcommand*\fsize{\dimexpr\f@size pt\relax}%
  \newcommand*\lineheight[1]{\fontsize{\fsize}{#1\fsize}\selectfont}%
  \ifx\svgwidth\undefined%
    \setlength{\unitlength}{5827.67468262bp}%
    \ifx\svgscale\undefined%
      \relax%
    \else%
      \setlength{\unitlength}{\unitlength * \real{\svgscale}}%
    \fi%
  \else%
    \setlength{\unitlength}{\svgwidth}%
  \fi%
  \global\let\svgwidth\undefined%
  \global\let\svgscale\undefined%
  \makeatother%
  \begin{picture}(1,0.26397277)%
    \lineheight{1}%
    \setlength\tabcolsep{0pt}%
    \put(0,0){\includegraphics[width=\unitlength,page=1]{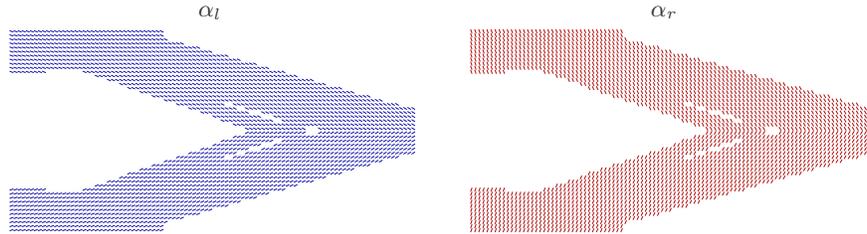}}%
    \put(0.22120044,0.25254099){\color[rgb]{0.2,0.2,0.2}\makebox(0,0)[lt]{\lineheight{1.25}\smash{\begin{tabular}[t]{l}$\alpha_{l}$\end{tabular}}}}%
    \put(0.74024266,0.25254099){\color[rgb]{0.2,0.2,0.2}\makebox(0,0)[lt]{\lineheight{1.25}\smash{\begin{tabular}[t]{l}$\alpha_{r}$\end{tabular}}}}%
  \end{picture}%
\endgroup%

    \caption{Fiber angle distributions for the best performing structure obtained with the BAxUS algorithm with the sequential strategy. The meanings of $\alpha_l$ and $\alpha_r$ are the same as in Fig. \ref{fig:svg_plot_1}. In this case $R_r=0.953$.}
    \label{fig:svg_plot_2}
\end{figure}

Figure~\ref{fig:svg_plot_2}, on the other hand, presents the fiber angle distribution corresponding to the best-performing configuration, obtained using the BAxUS algorithm with the sequential optimization strategy. In this case, the value of \( R_r \) is close to 1, indicating a stacking sequence of \( \pm[\alpha_r] \). Focusing on the distribution of \(\alpha_r\), the fiber orientations exhibit a distinct pattern: the fibers are oriented nearly vertically near the clamped boundary, while the orientation shifts toward approximately $\pm 60^{\circ}$ in the vicinity of the load application point. This configuration contrasts with the one shown in Figure~\ref{fig:svg_plot_1}.

By inspecting the lower half of $\alpha_r$ in Figure \ref{fig:svg_plot_2}, the fiber orientations are influenced by the specific configuration of the underlying MMC. Therefore, the idea from Sun~et~al.~\cite{sun_structural_2022} of having independent fiber orientation per MMC might be of use. This observation suggests the potential benefit of updating the benchmark by introducing additional master nodes whose position could be altered, as proposed by Serhat and Basdogan \cite{serhat_lamination_2019} to interpolate the LPs more effectively in response to the deformation of the underlying MMCs. 
% Such an approach could lead to fiber orientations that are better aligned with the structural response, thereby improving overall performance.

% Finally, as promised by the LPIM approach \cite{serhat_lamination_2019}, the distributions shown in Figures \ref{fig:svg_plot_2} and \ref{fig:svg_plot_1}, the orientation transition of the fibers is smooth.

\subsection{Comparing Solutions between Sequential and Concurrent Settings}
\begin{figure}[htbp]
    \centering
    % Placeholder box (optional)
	\includegraphics[width=0.9\linewidth,trim={0cm 0cm 3.65cm 2cm},clip]{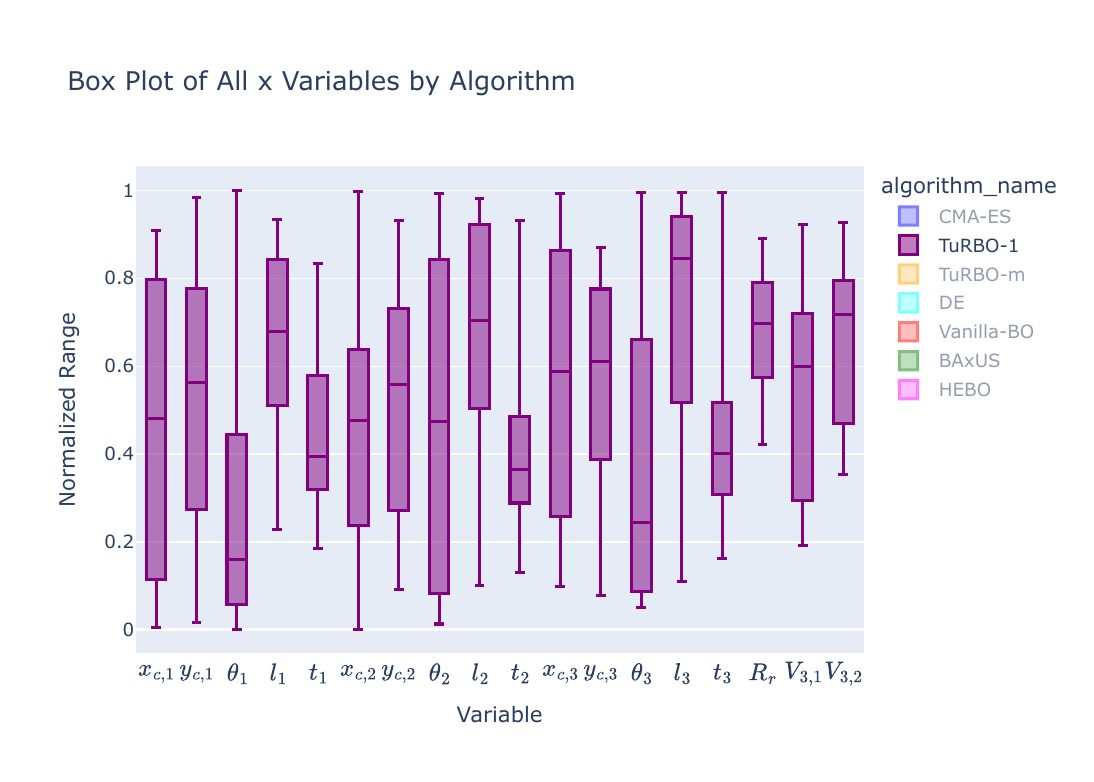}
    \caption{Boxplots demonstrating distribution of optimal design variables for TuRBO-1 algorithm and the concurrent approach.}
    \label{fig:box_plots_variable_turbo-1_concurrent}
\end{figure}

\begin{figure}[htbp]
    \centering
    % Placeholder box (optional)
	\includegraphics[width=0.9\linewidth,trim={0cm 0cm 3.65cm 2cm},clip]{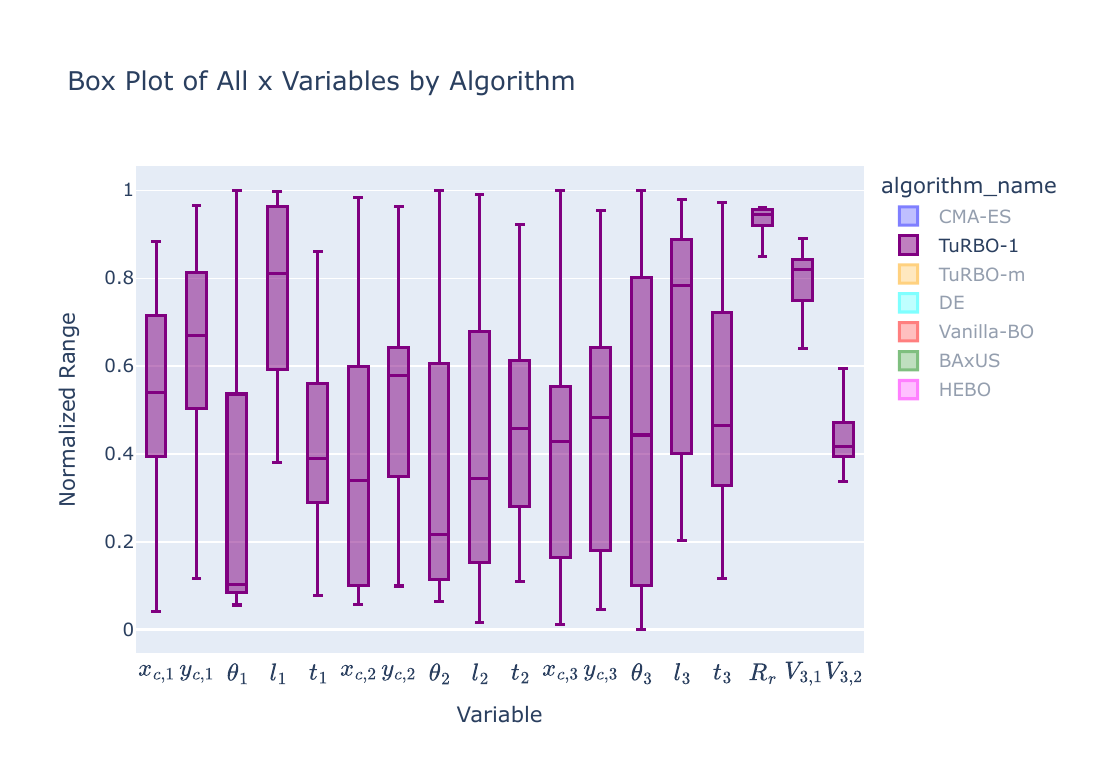}
    \caption{Boxplots demonstrating distribution of optimal design variables for TuRBO-1 algorithm and the sequential approach.}
    \label{fig:box_plots_variable_turbo-1_sequential}
\end{figure}

Upon examining Figure~\ref{fig:box_plots_compliance} with Tables~\ref{tab:concurrent_layouts} and~\ref{tab:sequential_layouts}, the optimization landscape appears highly multimodal, as reflected in the diverse topologies obtained. TuRBO-1 shows the best average performance, motivating a closer look at its variable behavior under the two optimization settings (Figures~\ref{fig:box_plots_variable_turbo-1_concurrent}-\ref{fig:box_plots_variable_turbo-1_sequential}, for the concurrent and sequential setting, respectively).
Across both the concurrent and sequential optimization settings, the variables associated with the three MMCs consistently span the entire admissible design space. This behavior indicates a highly multimodal objective landscape, characterized by a broad distribution of local optima. In contrast, the behavior of the lamination parameter vector, \(\mathbf{x}_{\text{LP}}\), reveals a more nuanced pattern. Under the concurrent formulation, the LPs display higher variability. Conversely, in the sequential approach, the optimal solutions exhibit a pronounced clustering in \(\mathbf{x}_{\text{LP}}\), suggesting a more targeted and effective exploitation of favorable LP-configurations. Although we present boxplots only for TuRBO-1 for clarity of presentation, the same trend was consistently observed across all optimization algorithms used in this study, with the exception of DE, which showed the lowest performance.

%This observation aligns with the discussion in Section~\ref{subsec:global_results}, where it was noted that trust-region methods can limit exploration in certain regions of the \(\mathbf{x}_{\text{LP}}\) subspace.

\section{Conclusions}
\label{sec:conclusions}
This work introduced a novel topology optimization framework for variable stiffness composite structures, using lamination parameter interpolation to enable smooth variations in fiber orientations throughout the design domain. By combining material layout (topology) and lamination parameter optimization, the proposed approach effectively enhances the stiffness of the structure.

A central aim of this study is to examine how the definition of the optimization problem, particularly the separation of topology and material variables, interacts with algorithmic choices. 
We study how the choice between a sequential and a concurrent formulation, along with the way the computational budget is allocated, affects both the quality and the realism of the final design. This perspective emphasizes that successful optimization in engineering design depends not only on algorithm selection but also on problem formulation and the incorporation of domain knowledge. When physical insights, such as differences in the nature of design variables or recognizable structures in the objective landscape, are available, they can be leveraged to guide the problem formulation instead of defaulting to a purely black box approach.

Future work will explore extensions of the proposed method by increasing the number of master nodes used for lamination parameter interpolation and incorporating alternative topology formulations. This investigation will allow a deeper insight into the interaction between material layout and fiber steering; two tightly coupled subproblems addressed in this study. Additionally, we aim to apply the framework to more realistic engineering scenarios, contributing to the development of richer and more representative benchmark problems for black-box optimization. We identify the lack of such test beds as a limitation in current benchmarking practices, which in turn affects how algorithm selection and configuration problem tasks are framed and evaluated.

\bibliographystyle{splncs04}
\bibliography{mybibliography}

\appendix 
\end{document}